\documentclass[10pt,journal]{IEEEtran}
\usepackage{amsmath,amssymb,amsfonts,amsthm}
\usepackage{graphicx}
\usepackage{booktabs}
\usepackage{array}
\usepackage{multirow}
\usepackage{url}
\usepackage{xcolor}
\usepackage{enumitem}
\usepackage{mathtools}
\usepackage{cite}
\usepackage{hyperref}
\hypersetup{colorlinks=true,linkcolor=blue,citecolor=blue,urlcolor=blue}
\usepackage{balance}
\usepackage{tikz}
\usetikzlibrary{arrows.meta,positioning,calc,fit}
\theoremstyle{definition}
\newtheorem{definition}{Definition}
\newtheorem{assumption}{Assumption}
\theoremstyle{plain}
\newtheorem{theorem}{Theorem}

\newtheorem{proposition}{Proposition}
\newtheorem{corollary}{Corollary}
\theoremstyle{remark}

\newcommand{\Pspace}{\mathcal P}
\newcommand{\X}{\mathcal X}
\newcommand{\D}{\mathcal D}
\newcommand{\E}{\mathcal E}
\newcommand{\C}{\mathcal C}
\newcommand{\V}{\mathcal V}
\newcommand{\A}{\mathcal A}
\newcommand{\T}{\mathcal T}

\newcommand{\R}{\mathcal R}
\newcommand{\dd}{\mathrm d}
\newcommand{\pdata}{p_{\rm data}}
\newcommand{\RFPR}{\mathrm{R\mbox{-}FPR}}

\newcommand{\Rad}{\mathfrak R}

\begin{document}

\title{Self-Consistent Generative Paths via Admissible Random Variational Transport}
\author{Lei Luo, Yingzhen Zhang, and Jian Yang 
\thanks{Lei Luo, Yingzhen Zhang, and Jian Yang are with PCA Lab, Key Lab of Intelligent Perception and Systems for High-Dimensional Information of Ministry of Education, School of Computer Science and Engineering, Nanjing University of Science and Technology, Nanjing, China. E-mail: \{cslluo, yingzhenzhang, csjyang\}@njust.edu.cn. }}
\maketitle

\begin{abstract}
Modern generative models construct paths from simple priors to data distributions. Diffusion models use many stochastic denoising steps, flow matching learns continuous transport fields, consistency and distillation methods seek one- or few-step maps, adversarial models match terminal distributions, and variational autoencoders generate through latent stochastic kernels. Existing unifying frameworks mainly specify how such paths are constructed: by velocity fields, stochastic interpolants, Markov generators, Wasserstein proximal score dynamics, or fixed-point samplers. This paper studies a different question: when is a generated probability path self-consistent?

We define a self-consistent generative path as a probability path that is invariant under admissible local random variational transport corrections. Given a path $\Gamma=\{\mu_t\}_{t\in[0,1]}$, a local correction is represented by an admissible random variational transport operator
\[
\V_{\theta,t,\omega}^{\Delta t}(\mu)=\arg\min_{\nu}\{\D_{\theta,t,\omega}(\nu,\mu)+\E_{\theta,t,\omega}(\nu)+\C_{\theta,t,\omega}(\nu,\mu)\},
\]
and the path is self-consistent when it is a random fixed point of the induced path operator, $\Gamma^*(\vartheta\omega)=\A_{\theta,\omega}(\Gamma^*(\omega))$. The formulation contains random regularized optimal-transport proximal steps as a structured instance, while allowing non-OT divergences, latent kernels, adversarial constraints, causal discrete kernels, and terminal one-step maps.

The theory yields a random fixed-point path residual (R-FPR), measuring the gap between the actual generated path and an admissible local correction. We prove well-posedness, random fixed-point existence and attraction, non-contractive existence under topological degree assumptions, residual-to-generation error bounds, finite-sample residual concentration, neural proxy perturbation bounds, and continuous-time limits. We also give a graded model dictionary showing how diffusion/score models, flow matching, rectified flow, consistency/one-step generators, VAE kernels, WGAN/GAN updates, autoregressive kernels, Schrodinger bridges, and energy-based samplers enter the same self-consistency calculus as exact instances, limits, degeneracies, or compatible embeddings. The resulting viewpoint turns endpoint matching into path self-consistency testing and provides a practical residual for diagnosing failures, regularizing training, and controlling adaptive sampling.
\end{abstract}


\section{Introduction}

A generative model is often evaluated by the quality of its endpoint distribution. In image generation, a sampler starts from a simple prior $\pi$ and aims to produce a terminal law close to the data law $\pdata$. This endpoint view is natural, but it hides a basic geometric issue: most modern generators do not only define an endpoint map, they define an entire path of probability laws from noise to data. A practical failure illustrates the issue. A diffusion sampler may look excellent at 50 function evaluations but become blurry or structurally distorted at 5 evaluations; a one-step student may reach the correct region while producing high-frequency artifacts; a GAN may produce sharp images while silently losing modes. In all three cases, the endpoint view is incomplete because the local journey from prior to data is unreliable.

Diffusion models follow a long stochastic denoising path \cite{ho2020ddpm,song2021score}. Flow matching defines a continuous path through a learned velocity field \cite{lipman2023flowmatching}. Stochastic interpolants bridge prescribed distributions through stochastic processes \cite{albergo2023stochastic}. Generator matching studies Markov generators that describe infinitesimal evolution \cite{holderrieth2025generator}. Consistency models and distribution matching distillation compress a long path into one or a few steps \cite{song2023consistency,yin2024dmd}. GANs match a terminal distribution through an adversarial game \cite{goodfellow2014gan,arjovsky2017wgan}, whereas VAEs generate through a latent stochastic kernel \cite{kingma2013vae}. These paradigms differ algorithmically, but they all implicitly or explicitly produce a probability path.

\begin{figure*}[t]
\centering
\includegraphics[width=0.88\textwidth]{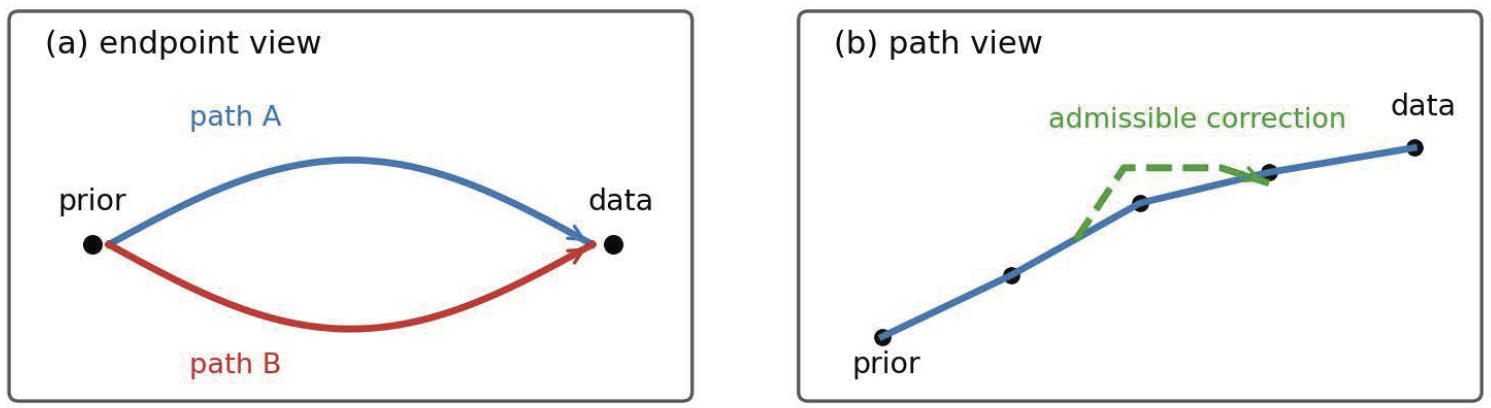}
\caption{Endpoint matching checks only the terminal law; path self-consistency requires each local transition to agree with an admissible correction.}
\label{fig:endpoint_path}
\end{figure*}

This paper asks a question that is complementary to path construction:
\begin{quote}
\emph{When is a generated probability path self-consistent?}
\end{quote}
Endpoint matching asks whether the destination looks right; path self-consistency asks whether the journey is reliable (Fig.~\ref{fig:endpoint_path}). Informally, a path is self-consistent if each local segment remains the same after an admissible variational correction. A path may reach a plausible endpoint while being locally unstable, excessively curved, over-smoothed, collapsed to a low-dimensional subset, or unsafe to compress into few steps. Such path-level defects appear in familiar forms: diffusion under-sampling, one-step distillation artifacts, GAN mode collapse, and VAE blur. We argue that these phenomena can be studied through a single principle: compare the model's actual local path step with an independent admissible variational correction of that step.

\subsection{Path self-consistency}
Let $\Gamma=\{\mu_t\}_{t\in[0,1]}$ denote a generated probability path with $\mu_0=\pi$ and $\mu_1$ intended to approximate $\pdata$. We introduce an admissible random variational transport operator (A-RVTO) that maps a current law $\mu_t$ to a locally corrected next law by solving or approximating a regularized variational transport step. A path is self-consistent if applying these local corrections reproduces the path. Formally,
\begin{equation}
\Gamma^*(\vartheta\omega)=\A_{\theta,\omega}(\Gamma^*(\omega)),
\label{eq:random_path_fp_intro}
\end{equation}
where $\omega$ is the random environment and $\vartheta$ is the base shift. Eq. \eqref{eq:random_path_fp_intro} is a random fixed-point equation on path space.

The important object is not a neural layer fixed point, not an individual sample-chain equilibrium, and not only an infinitesimal generator. It is the random fixed point of a corrected distribution path. This makes the framework deliberately different from fixed-point diffusion accelerators, Markov-process unifications, or Wasserstein proximal descriptions of score models.

\subsection{Why admissibility matters}
A naive residual could compare the model's actual next step with the same update rule applied again. This is degenerate: it primarily tests whether the sampler reproduces itself. To avoid this pitfall, the local correction used in the residual must be \emph{admissible}: it must not be identical to the actual sampler, and it must represent an independent or refined local variational correction such as a two-half-step reference, a higher-order solver, a teacher correction, a critic-induced descent step, or a latent-data product correction. This admissibility principle is central to the paper. It prevents the residual from collapsing to a numerical artifact and makes it a meaningful signal of path reliability.

\subsection{Contributions}
We make four contributions.
\begin{enumerate}[leftmargin=*,topsep=2pt,itemsep=2pt]
\item We formulate the \emph{path self-consistency problem} for generative modeling and define self-consistent generative paths as random fixed points of corrected probability paths.
\item We introduce \emph{admissible random variational transport operators} (A-RVTOs), with random regularized optimal-transport proximal steps (R-ROTOs) as a structured instance, and give an admissible-proxy principle that avoids same-sampler degeneracy.
\item We develop a theory of existence, attraction, non-contractive fixed points, residual-to-generation error control, empirical residual concentration, proxy perturbation, continuous-time limits, and operator-level generalization with model-specific corollaries.
\item We provide a graded model dictionary and an executable evaluation protocol showing how R-FPR can diagnose failures, regularize training, and control adaptive sampling across diffusion, flow, one-step, VAE, GAN/WGAN, and discrete/autoregressive generators.
\end{enumerate}

\subsection{Notation summary}
We use $\theta$ for model parameters and $\vartheta$ for the base shift of the random dynamical system $(\Omega,\mathcal F,\mathbb P,\vartheta)$. The exact local A-RVTO correction is $\V$, the actual sampler transition is $\T$, and the implementable admissible proxy is $\widetilde{\V}$. The path operator is $\A$, a probability path is $\Gamma=\{\mu_t\}_{t\in[0,1]}$, the population residual is $\RFPR(\Gamma)$, and its empirical estimate is $\widehat\RFPR(\Gamma)$. We reserve $\Pspace(\X)$ for probability laws on the state space and write $d_{\Pspace}$ for the chosen distribution metric.
R-FPR denotes the random fixed-point path residual.

\section{Related Work and Distinction}

\subsection{Path-based and process-based unifications}
Flow Matching trains continuous normalizing flows by regressing vector fields along conditional probability paths and includes diffusion paths as a special family \cite{lipman2023flowmatching}; generalized conditional flow matching further connects path design to minibatch optimal transport \cite{tong2023cfm}. Stochastic Interpolants unify flow and diffusion constructions through stochastic bridges between distributions \cite{albergo2023stochastic}. Generator Matching formulates generative modeling with arbitrary Markov processes and unifies diffusion, flow matching, and discrete diffusion through infinitesimal generators \cite{holderrieth2025generator}. These works primarily specify or learn dynamics. Our framework studies whether the generated path is invariant under admissible local variational corrections, and whether the resulting residual controls generation error.

\subsection{Wasserstein proximal views and OT generation}
The JKO scheme connects Fokker-Planck evolution to Wasserstein proximal steps \cite{jordan1998jko,ambrosio2005gradient,villani2009ot}. Recent Wasserstein proximal operator analyses describe score-based generative models through proximal dynamics and mean-field game optimality conditions \cite{zhang2024wpo}. Diffusion-transport methods such as DCNOT further combine diffusion cascades with neural optimal transport for multi-domain image-to-image translation \cite{zhang2025dcnot}. These works are an important foundation. Our R-ROTO instance includes such proximal and transport-based descriptions, but the present object is broader: a random path-space fixed point with admissible proxies, compatible embeddings beyond score models, and residual-control theory.

\subsection{Fast solvers and path discretization}
Fast diffusion samplers such as DPM-Solver and DEIS reduce sampling cost by exploiting the numerical structure of diffusion ODEs and by reducing local discretization error \cite{lu2022dpmsolver,zhang2023deis}. Related flow and diffusion editing methods, such as VRFNO and ERDDCI, improve path straightness or reversibility for few-step generation and high-quality editing \cite{dai2025vrfno,dai2025erddci}. These methods are closely related to our stepwise residual, but their goal is solver design, path straightening, or inversion. SCGP instead asks whether a local step is reliable relative to an admissible correction, so DPM-Solver, DEIS, Heun, two-half-step references, or teacher corrections can serve as admissible proxies. GAN convergence analyses also show that local dynamics can be unstable even when losses appear reasonable \cite{mescheder2018ganconverge}; our residual is a distribution-path diagnostic rather than a parameter-space convergence proof.
\subsection{Fixed points in generative modeling}
Deep equilibrium approaches to diffusion models formulate an augmented DDIM sampling chain as a multivariate fixed-point system \cite{pokle2022deqdiffusion}. Fixed Point Diffusion Models embed implicit fixed-point layers into diffusion denoisers \cite{bai2024fpdm}. A causal generative modeling work formulates structural causal models as fixed-point problems \cite{scetbon2024causal}. These methods demonstrate that fixed points are useful in generative modeling. The fixed point here is different: it lives on the generated distribution path, not primarily on a sample chain, a denoising module, or a causal variable system.

\subsection{One-step generation and terminal consistency}
Consistency models map noisy states to clean data and support one-step and few-step generation \cite{song2023consistency}. Distribution Matching Distillation converts diffusion models into one-step generators by distribution-level matching \cite{yin2024dmd}. Recent fast-generation work further emphasizes large-step path reliability: Shortcut Models condition on the desired step size \cite{frans2025shortcut}, MeanFlow learns average velocities for one-step generation \cite{geng2025meanflow}, SANA-Sprint uses continuous-time consistency distillation for one-step text-to-image generation \cite{chen2025sanasprint}, and DiMO distills masked diffusion models into one-step generators \cite{zhu2025dimo}. FastFlow also accelerates flow-matching pipelines without retraining \cite{bajpai2026fastflow}. In our terminology, these methods are large-step, shortcut, or terminal self-consistency mechanisms. We do not treat them as competitors; they supply important examples where path residuals can reveal when one-step or few-step compression is reliable.

\begin{table*}[t]
\centering
\caption{Comparison by primary mathematical object. The table is not a checklist of superiority; it clarifies that SCGP studies path-level self-consistency rather than only dynamics construction.}
\label{tab:related_objects}
\footnotesize
\begin{tabular}{p{2.35cm}p{2.8cm}p{3.0cm}p{1.85cm}p{4.45cm}}
\toprule
Framework & Primary object & Main question & Path-level residual? & Difference from SCGP/A-RVTO \\
\midrule
Flow Matching~\cite{lipman2023flowmatching} & Velocity field on prescribed probability paths & How to learn transport dynamics & No & Does not test the whole generated path as a fixed point of admissible local corrections \\
Stochastic Interpolants~\cite{albergo2023stochastic} & Stochastic bridge/interpolant & How to unify flows and diffusions & No & Bridge construction rather than residual-controlled path self-consistency \\
Generator Matching~\cite{holderrieth2025generator} & Infinitesimal Markov generator & How to generate with arbitrary Markov processes & No & Generator-level dynamics rather than variational correction residuals on distribution paths \\
WPO for score models~\cite{zhang2024wpo} & Wasserstein proximal score dynamics & What proximal structure explains score models & Score-specific & Deterministic score/diffusion-centered subcase of the R-ROTO instance \\
Fast solvers and samplers~\cite{lu2022dpmsolver,zhang2023deis,bajpai2026fastflow} & Numerical integration or adaptive inference & How to reduce NFE or wall-clock time & Usually no & R-FPR gives a model-agnostic residual for deciding where refinement is needed \\
Fixed-point diffusion~\cite{pokle2022deqdiffusion,bai2024fpdm} & Sample chain or denoising-module fixed point & How to accelerate or parameterize diffusion sampling & No & Fixed point is not the corrected distribution path \\
Fast one-step/few-step models~\cite{song2023consistency,yin2024dmd,frans2025shortcut,geng2025meanflow,chen2025sanasprint} & Terminal map, average velocity, shortcut, or distillation objective & How to compress generation into one or few steps & Terminal only & Enter SCGP as large-step terminal self-consistency problems \\
SCGP/A-RVTO & Corrected distribution path & Is the generated path reliable under admissible local correction? & Yes & Provides a path residual and residual-control theory \\
\bottomrule
\end{tabular}
\end{table*}

\section{Self-Consistent Generative Paths}

\subsection{Probability path space}
Let $(\X,d_\X)$ be a Polish state space, and let $\Pspace(\X)$ denote a set of probability measures on $\X$ equipped with a probability metric $d_{\Pspace}$. In most theoretical statements $d_{\Pspace}$ is either $W_2$, a bounded MMD metric induced by a characteristic kernel, a Sinkhorn-type metric satisfying stated stability assumptions, or a causal/discrete metric on sequence laws. Let $(\Omega,\mathcal F,\mathbb P,\vartheta)$ be a metric dynamical system representing stochastic environments such as diffusion noise, latent variables, minibatch randomness, critic fluctuation, prompts, or domain perturbations \cite{arnold1998random,crauel1992random}.

\begin{definition}[Probability path]
A probability path is a measurable curve $\Gamma:[0,1]\to \Pspace(\X)$, written $\Gamma(t)=\mu_t$. For a nonnegative weight $w$ with $\int_0^1 w(t)\,\dd t=1$, define
\begin{equation}
 d_\Gamma(\Gamma,\Lambda)=\left(\int_0^1 d_{\Pspace}^2(\mu_t,\nu_t)w(t)\,\dd t\right)^{1/2},
\end{equation}
where $\Lambda(t)=\nu_t$. Boundary conditions are $\mu_0=\pi$ and $\mu_1\in\mathcal N(\pdata)$, where $\pi$ is a simple prior and $\mathcal N(\pdata)$ is a prescribed neighborhood of the data law.
\end{definition}

The path metric lets us distinguish terminal quality from path quality. Two samplers may produce comparable endpoints but differ substantially in their intermediate marginals, local curvature, stochastic stability, or compressibility.

\subsection{Admissible random variational transport}
The mother operator used in this paper is not restricted to optimal transport. It is a random variational transport step.

\begin{definition}[A-RVTO step]
Given a law $\mu\in\Pspace(\X)$, a time $t$, a step size $\Delta t$, and a random environment $\omega$, an admissible random variational transport step is represented by
\begin{equation}
\begin{aligned}
\V_{\theta,t,\omega}^{\Delta t}(\mu)
\in \arg\min_{\nu\in\Pspace(\X)}
\big\{&\D_{\theta,t,\omega}^{\Delta t}(\nu,\mu)
+\E_{\theta,t,\omega}^{\Delta t}(\nu)\\
&+\C_{\theta,t,\omega}^{\Delta t}(\nu,\mu)\big\}.
\end{aligned}
\label{eq:arvto}
\end{equation}
Here $\D$ is a geometry or divergence term, $\E$ is a data/semantic/score/critic/ELBO energy, and $\C$ is a structural constraint such as generator-manifold, latent, causal, reversible, or multimodal consistency.
\end{definition}

\begin{definition}[R-ROTO instance]
A random regularized optimal-transport step is the special case
\begin{equation}
\R_{\theta,t,\omega}^{\Delta t}(\mu)
\in
\arg\min_{\nu}
\left\{
\frac{1}{2\Delta t}W_2^2(\nu,\mu)+\E_{\theta,t,\omega}^{\Delta t}(\nu)
\right\}.
\label{eq:rroto}
\end{equation}
We use R-ROTO as the main geometrically structured instance, while A-RVTO is the general framework.
\end{definition}

\subsection{Why A-RVTO is not an empty wrapper}
The variational form in \eqref{eq:arvto} is intentionally broad, but it is not used as an unfalsifiable reparameterization of arbitrary algorithms. The empirical object is the admissible residual, and admissibility imposes three testable restrictions: the correction cannot be the same sampler, it must be a refinement or independent variational correction, and it must pass pre-specified stress tests. Thus the theory can be falsified: if no admissible proxy separates normal and stressed paths, the claimed residual is rejected for that model class. This makes SCGP a residual-control principle rather than a notation-level taxonomy.

\subsection{Path operator and random fixed point}
For a grid $0=t_0<t_1<\cdots<t_K=1$, define a path correction operator $\A_{\theta,
\omega}$ by
\begin{equation}
\tilde\mu_{t_{k+1}}=
\V_{\theta,t_k,\vartheta^k\omega}^{\Delta t_k}(\tilde\mu_{t_k}),
\quad \tilde\mu_{t_0}=\mu_{t_0},
\label{eq:path_operator}
\end{equation}
and interpolate between grid points. Then $\A_{\theta,\omega}$ maps a candidate path $\Gamma$ to a corrected path $\tilde\Gamma$.

\begin{definition}[Self-consistent generative path]
A random path $\Gamma^*:\Omega\to\Pspace$ is self-consistent if
\begin{equation}
\Gamma^*(\vartheta\omega)=\A_{\theta,\omega}(\Gamma^*(\omega)) \quad \text{a.s.}
\label{eq:scgp}
\end{equation}
If additionally $\Gamma^*_1\in\mathcal N(\pdata)$, the path is data-calibrated.
\end{definition}

\noindent A good generator is not merely a map from noise to data. It is a path whose local admissible corrections reproduce the path itself. This viewpoint separates two questions that are often mixed: whether a trajectory is constructed, and whether that trajectory is self-consistent.

\section{Admissible Proxies and Same-Sampler Degeneracy}

Exact variational transport steps are rarely computable for modern neural generators. Experiments and algorithms therefore use a proxy $\widetilde{\V}_{\theta,t,\omega}^{\Delta t}$. The proxy is not an arbitrary implementation detail: if it equals the actual sampler, the residual becomes uninformative.

\subsection{Actual step versus correction step}
Let $\T_{\theta,t,\omega}^{\Delta t}$ denote the model's actual local path transition, so that an observed path pair is approximately $(\mu_t,\T_{\theta,t,\omega}^{\Delta t}\mu_t)$. The residual must compare the actual step with an \emph{admissible correction} $\widetilde{\V}$, not with the same transition $\T$.

\begin{definition}[Admissible local correction proxy]
A proxy $\widetilde{\V}_{\theta,t,\omega}^{\Delta t}$ is admissible relative to $\T_{\theta,t,\omega}^{\Delta t}$ if it satisfies the following conditions.
\begin{enumerate}[leftmargin=*,topsep=2pt,itemsep=2pt]
\item \emph{Non-identity:} $\widetilde{\V}_{\theta,t,\omega}^{\Delta t}$ is not identically the same as $\T_{\theta,t,\omega}^{\Delta t}$ on the evaluated path family.
\item \emph{Local refinement:} $\widetilde{\V}$ is a higher-order, teacher, two-half-step, energy-descent, critic-induced, or latent-data correction intended to be closer to the local variational step than the raw transition.
\item \emph{Proxy error:} there exists $\varepsilon_{\rm proxy}$ such that
\begin{equation}
 d_{\Pspace}(\widetilde{\V}_{\theta,t,\omega}^{\Delta t}\mu,
 \V_{\theta,t,\omega}^{\Delta t}\mu)
 \leq \varepsilon_{\rm proxy}(t,\omega,\mu)
\label{eq:proxy_error}
\end{equation}
for the evaluated path family.
\item \emph{Falsifiability:} under pre-specified stress perturbations of the generator or path discretization, the induced residual must increase or the proxy is rejected for the main experiment.
\end{enumerate}
\end{definition}

\begin{figure*}[t]
\centering
\includegraphics[width=0.8\textwidth]{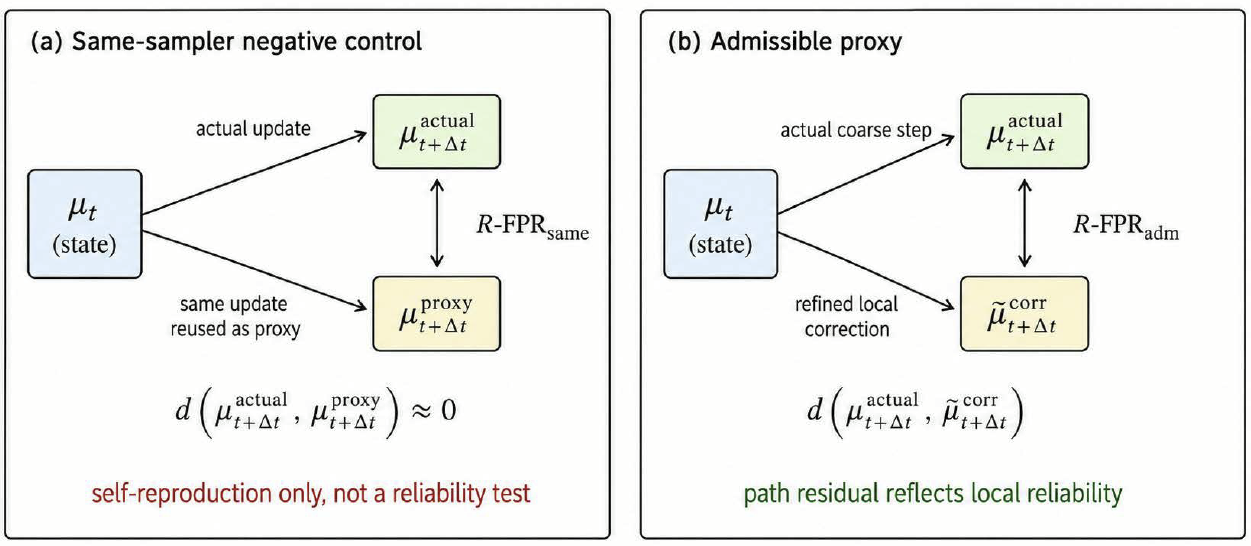}
\caption{Admissible proxy principle. The left panel is a same-sampler negative control: reusing the actual update as the proxy makes the residual artificially close to zero and only tests self-reproduction. The right panel is the main residual used in this paper: an admissible proxy provides an independent or refined local correction, so the resulting distance measures local path reliability.}\label{fig:proxy}
\end{figure*}

\begin{proposition}[Same-sampler degeneracy]
If $\widetilde{\V}_{\theta,t,\omega}^{\Delta t}=\T_{\theta,t,\omega}^{\Delta t}$ and the actual next marginal is sampled from $\T_{\theta,t,\omega}^{\Delta t}\mu_t$, then the population residual between the actual next marginal and the proxy next marginal is zero up to sampling error. Consequently, the residual cannot distinguish path self-consistency from sampler self-reproduction.
\end{proposition}
\noindent This proposition is a safeguard. Same-sampler residuals are useful as negative controls for implementation, but they cannot serve as the main self-consistency signal; Fig.~\ref{fig:proxy} summarizes the distinction.

\subsection{Model-specific admissible proxies}
Table \ref{tab:proxy} gives the proxies used by the evaluation protocol. Each has a corresponding negative control and a stress test.

\begin{table*}[t]
\centering
\caption{Admissible proxy choices. The proxy is deliberately stronger or independent relative to the actual transition.}
\label{tab:proxy}
\begin{tabular}{p{2.4cm}p{3.4cm}p{5.2cm}p{4.0cm}}
\toprule
Model family & Actual local step & Admissible correction proxy & Negative control \\
\midrule
Diffusion/EDM & Coarse reverse step or chosen sampler step & Two-half-step reference, Heun/DPM-Solver local correction, or teacher-score correction with the same schedule family & Same reverse step \\
Flow matching / rectified flow & Euler or chosen ODE solver step & Heun, midpoint, two-half-step reference, or curvature-aware correction & Same Euler step \\
Consistency / one-step & Student one-step or few-step map & Teacher few-step or refined terminal map & Same student map \\
GAN/WGAN & Generator endpoint transition & Critic-induced constrained descent followed by feature or generator-manifold projection & No critic correction \\
VAE / beta-VAE & Encoder-decoder reconstruction or latent traversal & Latent-data product correction combining reconstruction descent and latent prior correction & Same reconstruction kernel \\
AR/discrete & Token kernel transition & Causal KL or masked-token refinement with teacher/history correction & Same token sampler \\
\bottomrule
\end{tabular}
\end{table*}

\section{Theoretical Analysis}

This section states the main theoretical results. Proofs are deferred to the Appendix. The results are organized to show how local admissible corrections lead to path fixed points, how residuals control generation error, and how empirical estimates generalize. The central result is not merely that several generators can be written in a common notation, but that the proposed path residual enters a generation-error bound: under stability, calibration, admissible-proxy accuracy, and statistical concentration, terminal generation error is controlled by R-FPR together with calibration, proxy, discretization, and statistical errors.

\subsection{Well-posedness}
\begin{assumption}[Basic variational regularity]
For each $(\theta,t,\omega)$, the functional in \eqref{eq:arvto} is proper, bounded from below, lower semicontinuous in the chosen topology, and has compact sublevel sets on the evaluated path family. The mapping $(t,\omega,\mu)\mapsto \D+\E+\C$ is jointly measurable. When uniqueness is claimed, the objective is strictly convex along the chosen geodesic or Hilbert embedding.
\end{assumption}
The compact-sublevel condition is not meant to assert global compactness of an unrestricted neural generator. In applications it is verified on the evaluated path family, by bounded latent priors, finite feature embeddings, coercive regularizers, or compactification by small smoothing noise. In Wasserstein settings it can also be justified through the usual tightness and moment-control criteria.

\begin{theorem}[Existence and measurability of A-RVTO]
Under basic variational regularity, the minimizer set in \eqref{eq:arvto} is nonempty. If the objective is strictly convex along the relevant geometry, the minimizer is unique and $\V_{\theta,t,\omega}^{\Delta t}$ is a measurable random operator. The induced path correction operator $\A_{\theta,\omega}$ is measurable on path space.
\end{theorem}
\noindent The local correction is a legitimate random operator, not merely a formal notation. This is essential before asking for random fixed points.

\subsection{Contraction and random attraction}
\begin{assumption}[Random Lipschitz condition]
There exists a random variable $L(\omega)$ with $\mathbb E\log L(\omega)<0$ such that
\begin{equation}
 d_\Gamma(\A_{\theta,\omega}\Gamma,\A_{\theta,\omega}\Lambda)
 \leq L(\omega)d_\Gamma(\Gamma,\Lambda)
\label{eq:random_contraction}
\end{equation}
for all paths in the evaluated complete path space.
\end{assumption}

\begin{theorem}[Random self-consistent path]
Suppose the path space is complete and \eqref{eq:random_contraction} holds. Then there exists a unique random self-consistent path $\Gamma^*$ satisfying \eqref{eq:scgp}. Moreover, for every deterministic initial path $\Gamma_0$, the pullback iterates converge almost surely:
\begin{equation}
 d_\Gamma(\A_{\theta,\vartheta^{-1}\omega}\circ\cdots\circ \A_{\theta,\vartheta^{-n}\omega}\Gamma_0,\Gamma^*(\omega))\to 0.
\end{equation}
\end{theorem}
\noindent Stable generative paths are attracting random fixed points. The condition $\mathbb E\log L<0$ has a direct mechanism: local corrections may expand in some random environments, but the average logarithmic effect must be contractive. This condition characterizes stable regimes; it is not asserted to hold globally for every neural generator. When it fails, the non-contractive existence theorem below still supplies an existence-level statement, and empirical contraction defects are reported in experiments.

\begin{corollary}[Verifiable contraction for proximal geometry]
For the R-ROTO instance, if the energy is $\alpha_{t,\omega}$-geodesically convex in the EVI sense and the resolvent satisfies
\begin{equation}
 W_2(\R_{\theta,t,\omega}^{\Delta t}\mu,\R_{\theta,t,\omega}^{\Delta t}\nu)
 \leq \frac{1}{1+\Delta t\alpha_{t,\omega}}W_2(\mu,\nu),
\end{equation}
then the path Lipschitz constant is bounded by the product of the local constants. Positive average convexity implies $\mathbb E\log L<0$.
\end{corollary}
\noindent The random contraction condition is not purely abstract. In structured proximal regimes it follows from energy geometry.

\subsection{Non-contractive existence}
GANs, one-step maps, and strongly adversarial updates need not be contractive. The next theorem is not used for stability rates; it ensures existence under non-contractive topology.

\begin{assumption}[Topological degree regime]
Let $U(\omega)$ be a random bounded open subset of path space. The map $\A_{\theta,\omega}$ is random condensing with respect to a measure of noncompactness, $I-\A_{\theta,\omega}$ has no zero on $\partial U(\omega)$, and the random Leray-Schauder degree satisfies
\begin{equation}
 \deg(I-\A_{\theta,\omega},U(\omega),0)\neq 0.
\end{equation}
\end{assumption}

\begin{theorem}[Existence without contraction]
Under the topological degree regime, $\A_{\theta,\omega}$ has at least one random fixed path in $U(\omega)$.
\end{theorem}
\noindent The theory is not limited to contractive diffusion-like models. It separates existence of a self-consistent path from uniqueness and attraction. This is important for adversarial and large-step generators.

\subsection{R-FPR and residual-control theory}
\begin{definition}[Fixed-point path residual]
Given an admissible proxy $\widetilde{\A}_{\theta,\omega}$, define
\begin{equation}
\RFPR(\Gamma)=
\left(\mathbb E_\omega d_\Gamma^2(\Gamma,\widetilde{\A}_{\theta,\omega}\Gamma)\right)^{1/2}.
\label{eq:rfpr}
\end{equation}
On a grid, this is estimated by
\begin{equation}
\widehat{\RFPR}^{\,2}=\sum_{k=0}^{K-1} w_k
\widehat d_{\Pspace}^{\,2}
\left(\widehat\mu_{t_{k+1}},
\widetilde{\V}_{\theta,t_k,\omega_k}^{\Delta t_k}(\widehat\mu_{t_k})\right).
\end{equation}
Here $\widehat d_{\Pspace}$ denotes the empirical distance estimator used in the experiment, for example MMD, Sinkhorn divergence, or sliced Wasserstein distance in a fixed feature space. The estimator and feature encoder are fixed before validation or testing.
\end{definition}

\begin{theorem}[Residual-to-fixed-path bound]
If the exact path operator $\A$ is $L$-contractive with $L<1$ and the admissible proxy error is bounded by $\varepsilon_{\rm proxy}$, then
\begin{equation}
 d_\Gamma(\Gamma,\Gamma^*)
 \leq
 \frac{\RFPR(\Gamma)+\varepsilon_{\rm proxy}}{1-L}.
\label{eq:residual_bound}
\end{equation}
\end{theorem}
\noindent This is the first main mechanism: if the candidate path is close to its admissible correction and the correction is stable, then the path is close to the ideal self-consistent path.

\begin{assumption}[Calibration and approximation]
There exists an ideal calibrated operator $\A^*$ whose fixed path $\Gamma^*$ satisfies $\Gamma^*_1=\pdata$. The learned operator differs from the ideal one by an energy calibration error $\varepsilon_{\rm cal}$, the proxy error is $\varepsilon_{\rm proxy}$, and the path discretization error is $\varepsilon_{\rm disc}$. The terminal projection $\Gamma\mapsto \Gamma_1$ is Lipschitz.
\end{assumption}

\begin{theorem}[Residual-calibration-generation bound]
Under the preceding assumptions, the terminal law $\mu_{\theta,1}$ of a candidate path $\Gamma_\theta$ satisfies
\begin{align}
 d_{\Pspace}(\mu_{\theta,1},\pdata)
 &\leq C_1\frac{\RFPR(\Gamma_\theta)}{1-L}
 +C_2\varepsilon_{\rm cal} \notag\\
 &\quad +C_3\varepsilon_{\rm proxy}
 +C_4\varepsilon_{\rm disc}.
\label{eq:generation_bound}
\end{align}
\end{theorem}
\noindent R-FPR is not introduced as a heuristic score. Under calibration, stability, proxy accuracy, and discretization control, it is a component of an explicit generation-error bound. This theorem is the main bridge from path self-consistency to generative quality. The constants depend on terminal evaluation Lipschitz constants, perturbation stability of the calibrated operator, and the local geometry of the chosen divergence, but not on the sample size once the statistical term is separated.

\subsection{Empirical generalization of residuals}
Because distributions are observed through samples, the residual itself must generalize.

\begin{theorem}[MMD residual concentration]
Let $k$ be a bounded characteristic kernel with $0\leq k\leq B$. Suppose each grid marginal is estimated from $n$ samples and the grid has $K$ intervals. For the unbiased squared MMD residual estimator, with probability at least $1-\delta$,
\begin{equation}
\left|\widehat{\RFPR}^{\,2}-\RFPR^2\right|
\leq C B\sqrt{\frac{\log(2K/\delta)}{n}}+\varepsilon_{\rm proxy}+\varepsilon_{\rm feat}.
\label{eq:mmd_concentration}
\end{equation}
Here $\varepsilon_{\rm feat}$ accounts for the use of frozen feature embeddings.
\end{theorem}
\noindent The residual can be estimated with standard finite-sample control. The proxy term is an approximation bias: increasing sample size reduces statistical error, not proxy bias. This is why the experimental protocol uses fixed feature encoders, multiple distances, confidence intervals, and proxy sanity checks.

\begin{corollary}[Path-level generalization]
If training and validation path residuals are estimated with independent samples and the same admissible proxy, their difference is controlled by the sum of two concentration terms of the form \eqref{eq:mmd_concentration}. Consequently, validation R-FPR is a statistically meaningful model-selection signal provided that hyperparameters are selected before test evaluation.
\end{corollary}
\noindent This corollary prevents misuse. R-FPR should be used with validation splits and fixed proxy rules, not tuned on the test set.

\subsection{Continuous-time mechanism}
For the R-ROTO instance with $D=W_2$, the Euler-Lagrange condition of the proximal step yields a formal velocity field. Here $\delta\E_t/\delta\mu$ denotes the first variation of the energy functional with respect to the probability measure; in smooth absolutely continuous settings it plays the role of a potential whose spatial gradient drives transport.
\begin{equation}
 v_t(x)=-\nabla \frac{\delta \E_t}{\delta \mu}(x).
\end{equation}
With stochastic or entropic regularization, the marginal dynamics take the form
\begin{equation}
 \partial_t\mu_t=-\nabla\cdot(v_t\mu_t)+\frac12\nabla^2:(a_t\mu_t).
\label{eq:fp_limit}
\end{equation}
Flow matching corresponds to the zero-diffusion first-order limit; score/diffusion models correspond to transport-diffusion limits; rectified flow adds a low-curvature or straightness preference.

\begin{proposition}[Formal continuous-time limit]
Under standard consistency and stability assumptions on the discrete variational scheme, the piecewise-interpolated A-RVTO path converges, as $\max_k\Delta t_k\to 0$, to a weak solution of \eqref{eq:fp_limit}. If $a_t=0$, the limit reduces to a continuity equation; if $a_t$ is nonzero, the limit is a Fokker-Planck type equation.
\end{proposition}
\noindent This formal derivation should be read as a consistency calculation rather than a full variational-convergence theorem. The framework is not tied to discrete samplers. It explains why diffusion and flow-like models appear as different limits of the same path self-consistency calculus.

\section{Operator-Level Generalization and Model-Specific Corollaries}
\label{sec:operator_generalization}

\begin{figure}[t]
\centering
\includegraphics[width=0.65\linewidth]{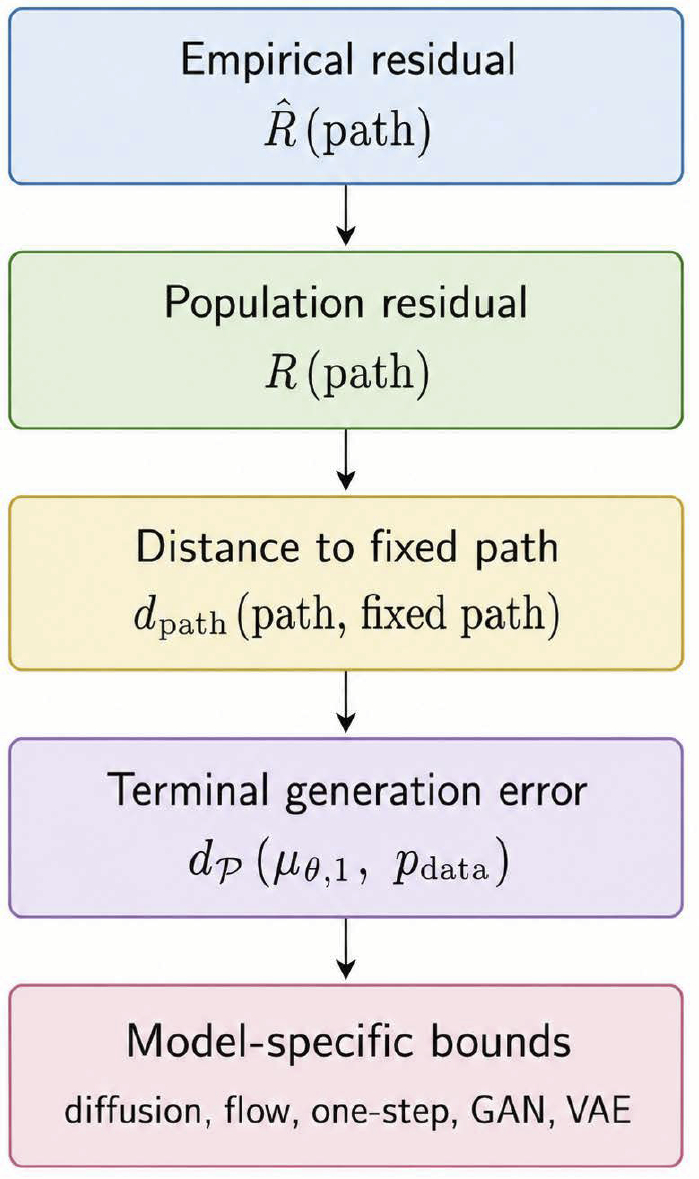}
\caption{Operator-level generalization. The empirical residual first generalizes to the population path residual. Under stability, this controls the distance to the self-consistent fixed path and hence the terminal generation error. Diffusion, flow, one-step, GAN, and VAE bounds are obtained by instantiating the same operator-level inequality with model-specific error terms.}\label{fig:generalization_ladder}
\end{figure}

The previous section shows that self-consistency has a fixed-point meaning. Fig.~\ref{fig:generalization_ladder} summarizes the resulting generalization chain. We now give a generalization analysis at the level of the A-RVTO operator. This is the correct level of analysis: individual models such as diffusion or flow matching inherit their bounds by instantiating the geometry, energy, constraint, randomness, and proxy class. The theory therefore has the following structure: a mother theorem for A-RVTO, followed by model-specific corollaries.

Let $Z$ collect all randomness used to estimate a stepwise residual: path samples, latent variables, sampler noise, random environments, and feature sampling. For each grid interval define the loss class
\begin{equation}
 g_{\theta,k}(Z)=d_{\Pspace}^2\left( X_{t_{k+1}},\widetilde{\V}_{\theta,t_k,\omega_k}^{\Delta t_k}(X_{t_k})\right),
\end{equation}
where $X_{t_{k+1}}$ is drawn from the actual path transition and the second argument is the admissible correction. Let $\mathcal G_k=\{g_{\theta,k}:\theta\in\Theta\}$.

\begin{theorem}[Operator-level uniform R-FPR generalization]
\label{thm:uniform_rfpr}
Assume $0\le g_{\theta,k}\le B$ for all $\theta,k$, and that the proxy family is fixed before test evaluation. Let $\widehat{\RFPR}^{\,2}(\theta)=\sum_{k=0}^{K-1}w_k n^{-1}\sum_{i=1}^n g_{\theta,k}(Z_{i,k})$ and let $\RFPR^2(\theta)=\sum_k w_k\mathbb E g_{\theta,k}(Z)$. Then with probability at least $1-\delta$, simultaneously for all $\theta\in\Theta$,
\begin{align}
\left|\widehat{\RFPR}^{\,2}(\theta)-\RFPR^2(\theta)\right|
&\le 2\sum_{k=0}^{K-1}w_k\Rad_n(\mathcal G_k) \notag\\
&\quad +B\sqrt{\frac{\log(2K/\delta)}{2n}}
+C\varepsilon_{\rm proxy}.
\label{eq:uniform_rfpr_rad}
\end{align}
If $\Theta$ has an $\epsilon$-covering number $\mathcal N(\epsilon,\Theta)$ and $g_{\theta,k}$ is $L_g$-Lipschitz in $\theta$, then the right hand side may be replaced by
\begin{align}
&CB\sqrt{\frac{\log \mathcal N(\epsilon,\Theta)+\log(2K/\delta)}{n}} \notag\\
&\qquad +C L_g\epsilon+C\varepsilon_{\rm proxy}.
\label{eq:uniform_rfpr_cover}
\end{align}
\end{theorem}
\noindent Empirical R-FPR is not merely a mini-batch quantity. Under standard complexity control, a low empirical residual implies a low population path residual. This theorem also explains why the experimental protocol must predefine the proxy class, feature encoder, grid, and distance family: otherwise the effective complexity term grows. For example, if $\Theta\subset\mathbb R^d$ is bounded and $g_{\theta,k}$ is $L_g$-Lipschitz, the covering number gives a rate of order $\sqrt{(d\log(L_g/\epsilon)+\log(K/\delta))/n}$ up to the approximation radius $\epsilon$. The term $\varepsilon_{\rm proxy}$ is an approximation bias of the chosen admissible correction; increasing sample size reduces statistical error but does not remove this proxy bias.

\begin{theorem}[Operator-level generation generalization]
\label{thm:generation_generalization}
Suppose the learned path operator is locally $L$-contractive around the calibrated path, $L<1$. Assume the ideal A-RVTO operator has a data-calibrated fixed path $\Gamma^*$ with $\Gamma^*_1=\pdata$, and let $\varepsilon_{\rm cal}(\theta)$, $\varepsilon_{\rm proxy}(\theta)$, $\varepsilon_{\rm disc}(\theta)$, and $\varepsilon_{\rm stat}(n,K,\Theta)$ denote energy calibration, proxy approximation, discretization, and statistical residual errors. Then, with probability at least $1-\delta$,
\begin{align}
 d_{\Pspace}(\mu_{\theta,1},\pdata)
 &\le C_1\frac{\widehat\RFPR(\Gamma_\theta)+\sqrt{\varepsilon_{\rm stat}(n,K,\Theta)}}{1-L} \notag\\
 &\quad +C_2\varepsilon_{\rm cal}(\theta)
 +C_3\varepsilon_{\rm proxy}(\theta)
 +C_4\varepsilon_{\rm disc}(\theta).
\label{eq:operator_generation_generalization}
\end{align}
\end{theorem}
\noindent This is the main generalization theorem. It states that terminal generation error is controlled by five interpretable quantities: empirical path residual, statistical complexity, calibration error, proxy error, and discretization error. In particular, reducing R-FPR is theoretically meaningful only when the proxy is admissible and calibration/discretization errors are not increased.

\begin{theorem}[Generalization of R-FPR regularized training]
\label{thm:regularized_training_gen}
Let $\hat\theta$ be an $\varepsilon_{\rm opt}$-approximate minimizer of
\begin{equation}
 \widehat{\mathcal L}_{\rm gen}(\theta)+\alpha\widehat\RFPR^{\,2}(\theta)
\end{equation}
over a fixed class $\Theta$. Suppose $\widehat{\mathcal L}_{\rm gen}$ and $\widehat\RFPR^{\,2}$ satisfy uniform convergence with complexity radii $\varepsilon_{\rm gen}(n)$ and $\varepsilon_{\rm rfpr}(n)$. Then
\begin{align}
 \mathcal L_{\rm gen}(\hat\theta)+\alpha\RFPR^2(\hat\theta)
 &\le \inf_{\theta\in\Theta}\{\mathcal L_{\rm gen}(\theta)+\alpha\RFPR^2(\theta)\} \notag\\
 &\quad +2\varepsilon_{\rm gen}(n)+2\alpha\varepsilon_{\rm rfpr}(n)+\varepsilon_{\rm opt}.
\label{eq:regularized_gen_bound}
\end{align}
\end{theorem}
\noindent R-FPR regularization is justified at the population level, not only at the training-batch level. The theorem does not claim that every quality metric must improve; rather, it identifies the conditions under which a lower empirical residual translates into a lower population self-consistency objective.

\subsection{Model-specific corollaries}
The following corollaries instantiate Theorem~\ref{thm:generation_generalization}. They are deliberately stated with model-specific error terms rather than forcing every model into the same exact proximal form.

\begin{corollary}[Diffusion/score models]
\label{cor:diffusion_gen}
Let $s_\theta(t,x)$ be a learned score and $s_*(t,x)$ the calibrated score. Suppose the admissible proxy has local order one higher than the actual sampler and the step size is $h$. Then
\begin{align}
 d_{\Pspace}(\mu_{\theta,0}^{\rm diff},\pdata)
 &\le C_1\frac{\widehat\RFPR_{\rm diff}+\sqrt{\varepsilon_{\rm stat}^{\rm diff}}}{1-L_{\rm diff}} \notag\\
 &\quad +C_2\varepsilon_{\rm score}(\theta)
 +C_3\varepsilon_{\rm proxy}^{\rm diff}+C_4h^p,
\label{eq:diffusion_cor_bound}
\end{align}
where $\varepsilon_{\rm score}(\theta)=\int\mathbb E\|s_\theta(t,x)-s_*(t,x)\|^2\,dt$ and $p$ is the local order of the sampler.
\end{corollary}
\noindent Diffusion error is controlled by path residual, score error, proxy accuracy, and sampling discretization. This explains why under-sampling raises stepwise R-FPR and why refined proxies are necessary.

\begin{corollary}[Flow matching and rectified flow]
\label{cor:flow_gen}
Let $v_\theta$ be the learned velocity field and $v_*$ the calibrated velocity. If the admissible proxy is a second-order local correction, then
\begin{align}
 d_{\Pspace}(\mu_{\theta,1}^{\rm flow},\pdata)
 &\le C_1\frac{\widehat\RFPR_{\rm flow}+\sqrt{\varepsilon_{\rm stat}^{\rm flow}}}{1-L_{\rm flow}} \notag\\
 &\quad +C_2\varepsilon_{\rm vel}(\theta)
 +C_3\varepsilon_{\rm proxy}^{\rm flow} \notag\\
 &\quad +C_4h^p+C_5\mathcal K_{\rm curv},
\label{eq:flow_cor_bound}
\end{align}
where $\varepsilon_{\rm vel}=\int\mathbb E\|v_\theta-v_*\|^2dt$. The curvature term is $\mathcal K_{\rm curv}=\int\mathbb E\|\partial_t v_\theta+D_xv_\theta v_\theta\|^2dt$.
\end{corollary}
\noindent Flow matching inherits a curvature-sensitive generalization form. Rectified or straighter flows reduce the curvature contribution, making few-step generation more reliable.

\begin{corollary}[Consistency and one-step generators]
\label{cor:onestep_gen}
Let $G_\theta$ be a student one-step generator and $T_{\rm teach}$ an admissible teacher few-step correction. Then
\begin{align}
 d_{\Pspace}((G_\theta)_\#\pi,\pdata)
 &\le C_1\left(\widehat\RFPR_{\rm one}+\sqrt{\varepsilon_{\rm stat}^{\rm one}}\right) \notag\\
 &\quad +C_2\varepsilon_{\rm teacher}
 +C_3\varepsilon_{\rm distill}+C_4\varepsilon_{\rm proxy}^{\rm one}.
\label{eq:onestep_cor_bound}
\end{align}
\end{corollary}
\noindent One-step quality is controlled by the student-teacher terminal residual, teacher quality, distillation error, and proxy accuracy. This is the formal reason R-FPR can predict one-step artifacts.

\begin{corollary}[GAN/WGAN constrained embedding]
\label{cor:gan_gen}
Let $\mathcal G=\{(G_\theta)_\#\pi:\theta\in\Theta\}$ be a generator family and let the critic-induced A-RVTO correction be restricted to $\mathcal G$. Under local stability,
\begin{align}
 d_{\Pspace}(\mu_{\theta}^{\rm GAN},\pdata)
 &\le C_1\frac{\widehat\RFPR_{\rm GAN}+\sqrt{\varepsilon_{\rm stat}^{\rm GAN}}}{1-L_{\rm GAN}} \notag\\
 &\quad +C_2\varepsilon_{\rm crit}
 +C_3\varepsilon_{\rm gen}+C_4\varepsilon_{\rm opt}.
\label{eq:gan_cor_bound}
\end{align}
\end{corollary}
\noindent GANs are not claimed to be exact unconstrained Wasserstein proximal models. Their bound contains critic calibration, generator-family approximation, and optimization terms. If the critic is updated $m$ times per generator step, the critic error is understood as $\varepsilon_{\rm crit}(m,n_{\rm batch},\mathcal D_{\rm critic})$, reflecting critic optimization, sample size, and function-class limitations. This precise wording prevents overclaiming while still yielding a useful failure diagnostic.

\begin{corollary}[VAE and beta-VAE]
\label{cor:vae_gen}
For a latent-data A-RVTO embedding with encoder $q_\phi(z|x)$ and decoder $p_\theta(x|z)$,
\begin{align}
 d_{\Pspace}(\mu_{\theta}^{\rm VAE},\pdata)
 &\le C_1\widehat\RFPR_{\rm VAE}+C_2\varepsilon_{\rm rec}
 +C_3\varepsilon_{\rm KL} \notag\\
 &\quad +C_4\varepsilon_{\rm post}+C_5\sqrt{\varepsilon_{\rm stat}^{\rm VAE}}.
\label{eq:vae_cor_bound}
\end{align}
\end{corollary}
\noindent The VAE residual separates data reconstruction and latent-prior mismatch. Increasing the KL weight can improve latent prior matching while worsening data-space sharpness, explaining blur as a path-level imbalance.

\begin{corollary}[Autoregressive and discrete generators]
\label{cor:ar_gen}
For a causal token-space A-RVTO with causal divergence $\D_{\rm causal}$ and history-conditioned proxy, terminal sequence-law error satisfies
\begin{align}
 d_{\rm causal}(\mu_{\theta}^{\rm AR},p_{\rm data})
 &\le C_1\widehat\RFPR_{\rm AR}+C_2\varepsilon_{\rm NLL}
 +C_3\varepsilon_{\rm hist} \notag\\
 &\quad +C_4\sqrt{\varepsilon_{\rm stat}^{\rm AR}}.
\end{align}
\end{corollary}
\noindent In discrete generation, self-consistency becomes a history-conditioned stability condition. The residual detects accumulation of local token-kernel errors.

\section{Mechanism and Generalization Analysis}

The preceding bounds justify R-FPR as a path-level residual. We next record several mechanisms that make the residual interpretable in concrete generators. These results are stated under regularity assumptions appropriate for the corresponding model class. Their role is not to claim that every generator satisfies all assumptions, but to clarify what R-FPR measures when the assumptions hold.

\subsection{Residual decomposition}
Let $\T_{\theta,t}^{\Delta t}$ be the actual local transition and $\widetilde{\V}_{\theta,t}^{\Delta t}$ an admissible correction. Introduce an ideal local correction $\V_{*,t}^{\Delta t}$ for the calibrated path. Then
\begin{align}
 d_{\Pspace}(\T_{\theta,t}^{\Delta t}\mu_t,\widetilde{\V}_{\theta,t}^{\Delta t}\mu_t)
 &\leq d_{\Pspace}(\T_{\theta,t}^{\Delta t}\mu_t,\V_{\theta,t}^{\Delta t}\mu_t) \notag\\
 &\quad + d_{\Pspace}(\V_{\theta,t}^{\Delta t}\mu_t,\V_{*,t}^{\Delta t}\mu_t) \notag\\
 &\quad + d_{\Pspace}(\V_{*,t}^{\Delta t}\mu_t,\widetilde{\V}_{\theta,t}^{\Delta t}\mu_t).
\label{eq:residual_decomp}
\end{align}
The three terms correspond respectively to actual-transition error, calibration error, and proxy approximation error.

\begin{proposition}[Residual decomposition]
Under Lipschitz stability of the local correction in the energy and in the input marginal, the stepwise residual is bounded by
\begin{align}
 r_t &\leq e_{\rm trans}(t)+C_{\rm cal}e_{\rm cal}(t)+e_{\rm proxy}(t) \notag\\
 &\quad +C_{\rm in}d_{\Pspace}(\mu_t,\mu_t^*).
\end{align}
Conversely, if the admissible proxy is a consistent local refinement and $e_{\rm proxy}$ is small, a large residual implies that at least one of transition mismatch, calibration mismatch, or path deviation is large.
\end{proposition}
\noindent R-FPR is not a mysterious scalar. It decomposes into identifiable mechanisms. This is why the protocol reports proxy sanity checks, calibration-sensitive stress tests, and stepwise residual curves rather than only one aggregate score.

\subsection{Flow and ODE mechanism}
For deterministic flow models, the most transparent admissible proxy is a higher-order local solver. Let the actual transition be the Euler step $x^E_{t+h}=x_t+h v(t,x_t)$ and the proxy be the two-half-step or midpoint/Heun step.

\begin{proposition}[Flow residual detects local curvature]
Assume $v$ is twice continuously differentiable with bounded derivatives. For an Euler actual step and a second-order admissible proxy, the local sample discrepancy satisfies
\begin{align}
 \|x^E_{t+h}-\widehat x_{t+h}\|
 &= \frac{h^2}{2}\left\|\partial_t v(t,x_t)
 +D_xv(t,x_t)v(t,x_t)\right\| \notag\\
 &\quad +O(h^3).
\label{eq:flow_curvature}
\end{align}
After averaging over $x_t\sim\mu_t$, the flow R-FPR estimates a path-curvature and local truncation signal.
\end{proposition}
\noindent For flow matching and rectified flow, R-FPR is not simply a new loss. It measures whether the learned vector field supports reliable local path propagation. Straight or low-curvature paths naturally have smaller residual and are easier to compress.

\subsection{Diffusion mechanism}
For diffusion samplers, the actual transition often uses a coarse discretization of a reverse SDE or probability-flow ODE. The admissible proxy can be a two-half-step, a higher-order solver, or a teacher-score correction.

\begin{proposition}[Diffusion residual detects under-sampling]
Suppose the reverse drift $b_\theta(t,x)$ and diffusion coefficient $\sigma(t)$ are Lipschitz, and the proxy has one higher local weak order than the actual step. Then the stepwise residual is bounded by
\begin{equation}
 r_t \leq C h^{p+1}+C h\,\mathbb E_{\mu_t}\|s_\theta(t,x)-s_*(t,x)\|+\varepsilon_{\rm proxy},
\label{eq:diffusion_residual}
\end{equation}
where $p$ is the weak order of the actual step and $s_*$ is the calibrated score. When score error is fixed, increasing the step size raises the residual at the expected local-discretization rate.
\end{proposition}
\noindent This explains why R-FPR can guide adaptive sampling: intervals with small residual can be merged, while intervals with large residual should be refined.

\subsection{Adversarial and variational mechanisms}
GANs and VAEs enter as compatible embeddings. For these models the residual should be interpreted through constraints and kernels rather than exact unconstrained $W_2$ proximality.

\begin{proposition}[GAN constrained residual]
Let $\mathcal G$ be a generator family and $\V_\mathcal G$ the constrained critic-induced correction over $\mathcal G$. If the implemented generator update has optimization error $\varepsilon_{\rm opt}$ and approximation error $\varepsilon_{\rm gen}$ relative to $\V_\mathcal G$, then
\begin{equation}
 d_{\Pspace}(\mu_{k+1},\V_\mathcal G\mu_k)
 \leq \varepsilon_{\rm opt}+\varepsilon_{\rm gen}.
\end{equation}
If coverage collapses to a subset on which the critic correction points outward, the constrained residual increases unless the generator family cannot represent the correction direction.
\end{proposition}
\noindent The GAN residual is a constrained diagnostic, not an exact OT identity. It is designed to reveal whether the generator endpoint path resists critic-induced correction.

\begin{proposition}[VAE latent-data residual]
On the product space $\X\times\mathcal Z$, let the latent divergence be weighted by $\beta$. If the decoder family is Lipschitz and the latent prior correction dominates the reconstruction correction as $\beta$ increases, then the data-space residual decomposes as
\begin{equation}
 r_{\rm VAE}\leq C_x r_{\rm rec}+C_z \beta r_{\rm prior}+\varepsilon_{\rm dec}.
\end{equation}
Large $\beta$ may reduce latent mismatch while increasing data-space blur or terminal mismatch.
\end{proposition}
\noindent This gives a path-level interpretation of beta-VAE blur: strong latent regularity can produce an entropy-dominated or prior-dominated self-consistency that is not data-sharp.

\subsection{Uniform generalization and model selection}
A common failure mode of new diagnostics is over-tuning. The following result states a simple uniform version of the concentration bound.

\begin{theorem}[Uniform validation bound over a finite protocol]
Consider a finite set $\mathcal H$ of model checkpoints, residual distances, feature encoders, and admissible proxies fixed before test evaluation. Under the bounded-kernel assumptions of the MMD residual concentration theorem, with probability at least $1-\delta$,
\begin{align}
\sup_{h\in\mathcal H}\left|\widehat{\RFPR}_h^2-\RFPR_h^2\right|
&\leq C B\sqrt{\frac{\log(2|\mathcal H|K/\delta)}{n}} \notag\\
&\quad +\varepsilon_{\rm proxy}+\varepsilon_{\rm feat}.
\label{eq:uniform_bound}
\end{align}
\end{theorem}
\noindent The experiment must pre-register the evaluated proxy families and feature metrics. The more variants are tried, the larger the validation uncertainty. This theorem motivates reporting all proxy and metric ablations rather than selecting a single favorable curve.

\begin{corollary}[Residual regularization improvement]
Assume the constants in \eqref{eq:generation_bound} are unchanged between a baseline path $\Gamma_b$ and a regularized path $\Gamma_r$. If $\RFPR(\Gamma_r)\leq \RFPR(\Gamma_b)-\Delta$ for $\Delta>0$, then the generation-error upper bound improves by at least $C_1\Delta/(1-L)$, up to changes in calibration, proxy, and discretization errors.
\end{corollary}
\noindent The theory does not promise that every metric always improves after residual regularization. It says what must be checked: residual reduction is meaningful only if calibration, proxy quality, and discretization are not worsened.

\begin{corollary}[Domain-shift sensitivity]
Suppose $\A_{\theta,\omega}$ is Lipschitz in the environment variable in an integral probability metric $d_\Omega$. If the environment distribution shifts from $P$ to $Q$, then
\begin{equation}
 |\RFPR_P(\Gamma)-\RFPR_Q(\Gamma)|
 \leq C_\Omega d_\Omega(P,Q)+\varepsilon_{\rm est}.
\end{equation}
\end{corollary}
\noindent R-FPR can be used as a stress-test signal under domain or corruption shifts, but changes in residual should be interpreted relative to the magnitude of the environment shift.

\section{Model Dictionary}

The claim is graded rather than literal. Table~\ref{tab:model_dictionary} gives the model dictionary. A model class may enter the calculus as an exact instance, a continuous-time limit, a large-step degeneracy, or a compatible embedding. This avoids the overclaim that every model is exactly the same proximal problem.

\begin{table*}[t]
\centering
\caption{Graded dictionary from model families to A-RVTO/SCGP.}
\label{tab:model_dictionary}
\begin{tabular}{p{2.3cm}p{2.6cm}p{4.3cm}p{5.2cm}}
\toprule
Model family & Inclusion level & A-RVTO interpretation & Mechanism explained \\
\midrule
Diffusion / score & Exact or proximal limit & Stochastic R-ROTO / WPO / JKO step with score/free-energy term & Sampling stability and under-sampling residual \\
Flow matching / CNF & Continuous-time limit & Zero-diffusion variational transport with velocity-induced energy & Path curvature and solver sensitivity \\
Rectified flow & Limit plus constraint & Low-curvature or straightness-constrained path & Why straight paths support fewer steps \\
Consistency / one-step / shortcut / MeanFlow & Large-step degeneracy & Terminal or average-velocity self-consistency with teacher/refinement proxy & Student-teacher gap, shortcut error, and one-step artifacts \\
VAE / beta-VAE & Kernelized embedding & Latent-data product RVTO with reconstruction and prior terms & Blur as excessive latent/entropy regularization \\
GAN / WGAN & Constrained embedding & Critic-energy RVTO over generator family & Collapse as non-contractive or low-coverage fixed set \\
Autoregressive / discrete & Causal extension & Causal KL or token-space variational transport & Error accumulation and history consistency \\
Schrodinger bridge & Entropic OT instance & Entropy-regularized bridge RVTO & Stochastic bridge as regularized path \\
EBM / Langevin & Energy-descent instance & Stochastic energy correction kernel & Mixing and local energy descent \\
\bottomrule
\end{tabular}
\end{table*}

\subsection{Diffusion and score models}
Reverse-time score dynamics induce marginal Fokker-Planck equations \cite{song2021score}. JKO theory and recent WPO analyses connect these evolutions to Wasserstein proximal structure \cite{jordan1998jko,zhang2024wpo}. Thus diffusion/score models are the most direct R-ROTO instances. In this case, R-FPR measures the discrepancy between the actual discretized reverse path and an admissible refined local proximal correction.

\subsection{Flow matching and rectified flow}
Flow matching learns $v_\theta(t,x)$ satisfying a continuity equation. This is the zero-diffusion limit of the A-RVTO dynamics. The admissible proxy should be a higher-order or two-half-step correction, not the same Euler step. R-FPR then measures local curvature, velocity inconsistency, and discretization sensitivity.

\subsection{Consistency and one-step generation}
One-step and shortcut-style models compress a path into a large terminal map or learn an average transition over a large interval. SCGP treats them as large-step degeneracies of path self-consistency. The admissible proxy is naturally a teacher few-step correction, a consistency-refined terminal map, or a higher-order reference transition. A large terminal R-FPR indicates that the student, shortcut, or average-velocity map is not faithful to the refined path.

\subsection{VAE and beta-VAE}
A VAE defines a Markov kernel $K_{\theta,\phi}(x,\dd y)=\int p_\theta(\dd y|z)q_\phi(\dd z|x)$. This is not an exact unconstrained $W_2$ proximal map. We treat it as a latent-data RVTO embedding: the reconstruction term acts in data space while the prior/KL term acts in latent geometry. This explains why strong latent regularization may produce an overly smooth terminal path.

\subsection{GAN and WGAN}
A WGAN critic defines a distributional potential. A constrained RVTO step over a generator family $\mathcal G=\{(G_\theta)_\#\pi\}$ is
\begin{equation}
\arg\min_{\nu\in\mathcal G}
\left\{\D(\nu,\mu_k)-\int D_\psi(x)\,\dd\nu(x)+\lambda\Omega(D_\psi)\right\}.
\end{equation}
This is a compatible embedding, not an exact unconstrained proximal identity. R-FPR is used to diagnose critic-induced correction gaps and mode coverage, not to claim exact equivalence.

\subsection{Autoregressive and discrete models}
Autoregressive models define causal kernels $p_\theta(x_t|x_{<t})$. With causal KL, Hamming-Wasserstein, or token-space metrics, these kernels become causal RVTO extensions. The residual measures whether local token transitions remain consistent with corrected history-conditioned distributions.

\section{R-FPR as a Method}

The practical residual compares the actual next marginal to an admissible correction marginal. It plays the role of a bridge from theory to implementation. The A-RVTO operator specifies the ideal local correction, the admissible proxy makes this correction computable, and R-FPR measures the gap between the actual path and the corrected path. The same quantity then has three practical uses: it diagnoses unreliable path segments, supplies a training regularizer, and controls adaptive sampling. The theoretical connection is given by the generation-error bound, where the residual appears together with calibration, proxy, discretization, and statistical errors.

\subsection{Diagnosis}
A high R-FPR indicates that the actual path deviates from its local correction. In diffusion this can indicate under-sampling; in flows, excessive curvature or solver error; in GANs, mode collapse or critic instability; in VAEs, over-regularized blur; in one-step models, student-teacher mismatch.

\subsection{Training regularization}
Given a baseline objective $\mathcal L_{\rm gen}$, one may train with
\begin{equation}
\mathcal L_{\rm total}=\mathcal L_{\rm gen}+\alpha\widehat{\RFPR}^{\,2}+\beta\mathcal C_{\rm contr},
\end{equation}
where $\mathcal C_{\rm contr}$ is an empirical contraction-defect penalty. The residual term is used only with an admissible proxy; same-sampler proxy is never used for the main objective.

\subsection{Adaptive sampling}
During diffusion or flow sampling, stepwise residuals can guide adaptive discretization. Low residual intervals may be skipped or merged; high residual intervals require refinement. This converts path self-consistency into an NFE-quality control signal.

\section{Experimental Protocol for Final Validation}

The empirical part will test falsifiable claims rather than only report quality metrics. The accompanying protocol specifies exact scripts and reporting rules. The central hypotheses are:
\begin{itemize}[leftmargin=*,topsep=2pt,itemsep=2pt]
\item H1: admissible R-FPR predicts FID/KID/Recall/Coverage and failure labels better than training loss alone.
\item H2: same-sampler proxy produces near-zero residual and is therefore a negative control, while admissible proxies separate normal and stressed paths.
\item H3: R-FPR regularization improves stability, coverage, or quality under matched compute.
\item H4: R-FPR adaptive sampling reduces NFE at fixed quality or improves quality at fixed NFE.
\end{itemize}

The required experiments are: proxy sanity checks, two-dimensional controlled validation, CIFAR-10 cross-model diagnosis, training regularization on at least three model families, adaptive sampling for diffusion/flow, and a high-resolution or application stress test. All metrics must be reported with at least five seeds, 95\% bootstrap confidence intervals, fixed validation-based hyperparameter selection, and multiple residual distances or feature spaces.
%\noindent\textbf{Why this protocol is necessary.} 
Without proxy sanity checks, R-FPR may degenerate into sampler self-reproduction. Without matched compute, regularization improvements may be unfair. Without multiple distances and confidence intervals, residual correlations may be feature-specific. These rules are part of the method rather than optional implementation details. 
The present version focuses on the theoretical framework and the validation protocol; full experimental results, including quantitative comparisons, ablation studies, and visual examples, will be included in the next version.

%\section{Design Implications}
\section{Design Implications}
%\section{Design Calculus and Limitations}

A-RVTO also provides a design coordinate system:
\begin{equation}
(\D,\E,\C,\omega,\Delta t,\Gamma).
\end{equation}
Changing $\D$ changes geometry; changing $\E$ changes semantic or physical goals; changing $\C$ changes structure; changing $\omega$ changes stochastic environment; changing $\Delta t$ changes many-step, few-step, or one-step generation; changing $\Gamma$ changes the domain from images to videos, 3D objects, multimodal paths, or discrete sequences.

\begin{figure}[t]
\centering
\includegraphics[width=0.85\linewidth]{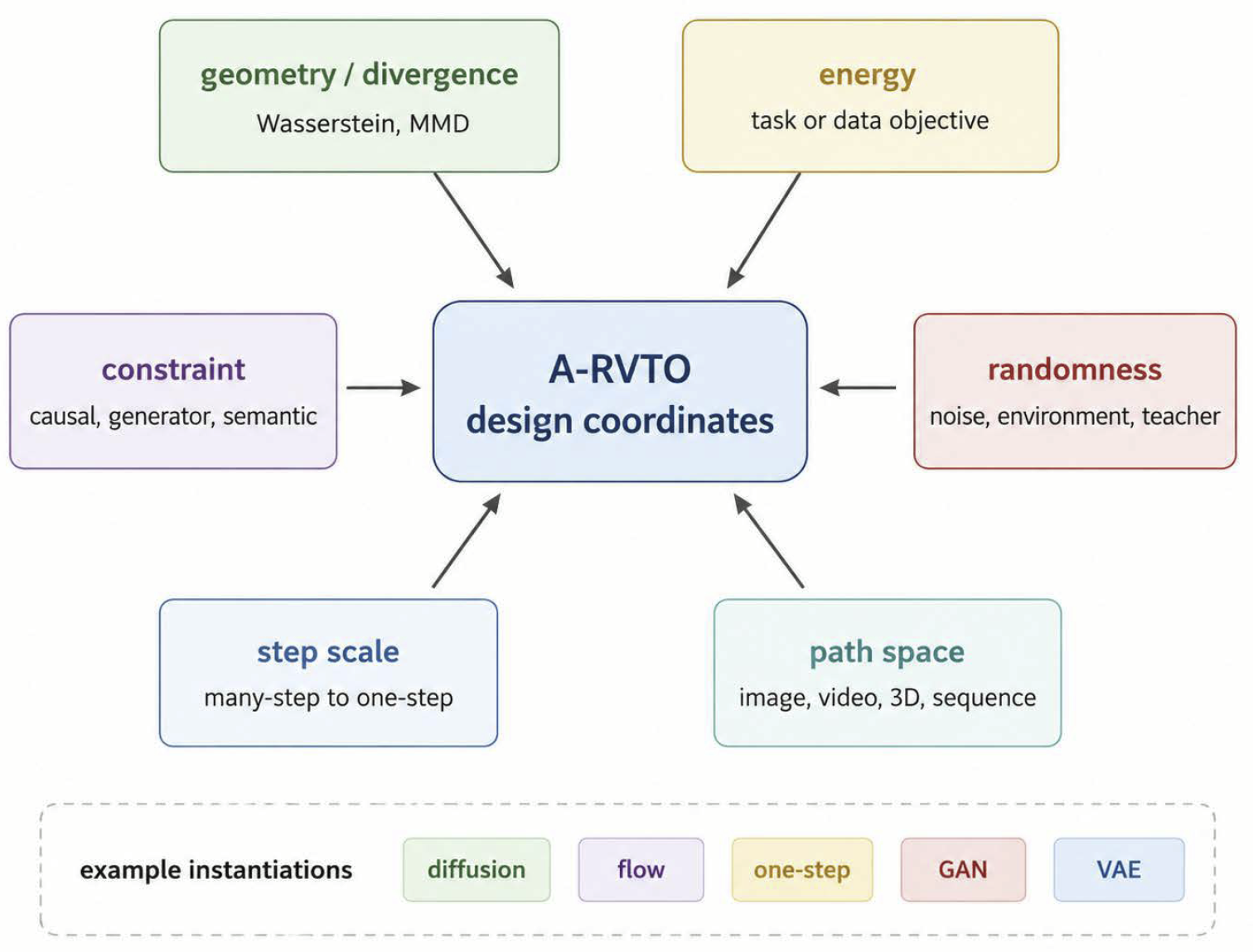}
\caption{Design coordinates of A-RVTO. A path-based generator is specified by the geometry or divergence, the energy, structural constraints, randomness, step scale, and path space. Diffusion, flow, one-step, GAN, and VAE models correspond to different choices of these coordinates rather than to unrelated principles.}\label{fig:design}
\end{figure}

The framework should not be interpreted as proving that every possible generator in mathematics is an A-RVTO model. The strong form applies to probabilistic generators admitting measurable path evolution, admissible local correction operators, and an attracting or invariant path. Pure symbolic program generators, non-measure-theoretic rule systems, or chaotic interactive generators without invariant laws may not naturally fit the strong framework. GAN, VAE, and autoregressive models are compatible embeddings rather than exact unconstrained Wasserstein proximal instances. These distinctions are intentional and prevent overclaiming.

\section{Conclusion}

We introduced self-consistent generative paths, a path-level fixed-point view of generative modeling. The central question is not merely whether a model reaches a realistic endpoint, but whether the journey from prior to data is locally reproducible under admissible random variational transport corrections. The A-RVTO framework, with R-ROTO as its optimal-transport instance, yields a fixed-point path residual that is theoretically connected to generation error and empirically testable as a diagnostic, regularizer, and adaptive sampling signal. The main conceptual message is simple: endpoint matching asks whether the destination looks right; path self-consistency asks whether the journey is reliable.
%We introduced self-consistent generative paths, a path-level fixed-point view of generative modeling. The central question is not only whether a generator reaches a realistic endpoint, but whether the journey from prior to data is locally reproducible under admissible random variational transport corrections. The A-RVTO framework, with R-ROTO as its optimal-transport instance, yields a random fixed-point path residual that is connected to generation error through residual-control and generalization bounds. The main conceptual message is that endpoint matching evaluates the destination, whereas path self-consistency evaluates the reliability of the journey. This viewpoint provides a common language for analyzing diffusion, flow, one-step, VAE, GAN/WGAN, and autoregressive generators as exact instances, limits, degeneracies, or compatible embeddings of a broader path-correction calculus. Future work will instantiate the residual in large-scale model families and study its empirical role in failure diagnosis, training regularization, and adaptive sampling.

\appendices
\section{Proofs and Additional Derivations}

\noindent\textbf{Reading guide.} The appendix follows the order of the statements in the main text. Each theorem, proposition, and corollary is restated by name through its subsection heading so that the proof can be located directly.

\subsection*{Proof of Proposition 1 (Same-sampler degeneracy)}
Assume that the proxy step equals the actual transition operator, namely
\[
\widetilde{\V}_{\theta,t,\omega}^{\Delta t}=\T_{\theta,t,\omega}^{\Delta t},
\qquad
\mu_{t+\Delta t}=\T_{\theta,t,\omega}^{\Delta t}\mu_t.
\]
Then the population discrepancy entering the residual is
\[
 d_{\Pspace}\!
 \left(
 \mu_{t+\Delta t},
 \widetilde{\V}_{\theta,t,\omega}^{\Delta t}(\mu_t)
 \right)
 = d_{\Pspace}\!\left(
 \T_{\theta,t,\omega}^{\Delta t}\mu_t,
 \T_{\theta,t,\omega}^{\Delta t}\mu_t
 \right)=0.
\]
The only nonzero contribution in practice therefore comes from finite-sample estimation, numerical approximation, or feature noise. Hence the quantity cannot reveal whether the path is genuinely reliable; it only reveals that the sampler can reproduce the same update rule. This proves the proposition.

\subsection*{Proof of Theorem 1 (Existence and measurability of A-RVTO)}
Let
\[
J_{\theta,t,\omega}^{\Delta t}(\nu;\mu)
=\D(\nu,\mu)+\E_{\theta,t,\omega}(\nu)+\C_{\theta,t,\omega}(\nu,\mu)
\]
denote the A-RVTO objective. By assumption, $J$ is proper and bounded from below, so there exists a minimizing sequence $\{\nu_n\}$. Because the sublevel sets of $J$ are compact on the evaluated path family, there is a subsequence $\nu_{n_j}$ converging in the chosen topology to some $\bar\nu$. Lower semicontinuity of $J$ yields
\[
J(\bar\nu;\mu)
\le \liminf_{j\to\infty}J(\nu_{n_j};\mu)
=\inf_{\nu}J(\nu;\mu),
\]
so $\bar\nu$ is a minimizer.

If the objective is strictly convex along the relevant geodesics or in the chosen Hilbert embedding, the minimizer is unique. Measurability of the minimizer multifunction follows from joint measurability of the data and the measurable maximum theorem. Under uniqueness, the measurable selection is single-valued, hence
\[
(\theta,t,\omega,\mu)\longmapsto \V_{\theta,t,\omega}^{\Delta t}(\mu)
\]
is a measurable random operator. The path operator $\A_{\theta,\omega}$ is a composition of finitely many measurable local correction maps on the discretized grid, so it is measurable on path space. This proves the theorem.

\subsection*{Proof of Theorem 2 (Random self-consistent path)}
For each $\omega$, define the pullback composition
\[
\A_{\omega}^{(n)}
=\A_{\theta,\vartheta^{-1}\omega}
\circ\cdots\circ
\A_{\theta,\vartheta^{-n}\omega}.
\]
If $\Gamma$ and $\Lambda$ are two candidate paths, repeated use of the random Lipschitz condition gives
\[
 d_\Gamma\!\left(\A_{\omega}^{(n)}\Gamma,\A_{\omega}^{(n)}\Lambda\right)
 \le \Bigg(\prod_{i=1}^{n}L(\vartheta^{-i}\omega)\Bigg)d_\Gamma(\Gamma,\Lambda).
\]
Since $\mathbb E\log L<0$, Birkhoff's ergodic theorem implies
\[
\frac1n\sum_{i=1}^{n}\log L(\vartheta^{-i}\omega)
\to \mathbb E\log L<0
\quad\text{a.s.}
\]
Hence the product $\prod_{i=1}^{n}L(\vartheta^{-i}\omega)$ converges almost surely to zero at an exponential rate.

Fix any deterministic initial path $\Gamma_0$. The exponential contraction estimate implies that the distance between two sufficiently long pullback iterates is bounded by a tail of exponentially decaying factors, multiplied by the usual tempered one-step increments. Hence $\{\A_{\omega}^{(n)}\Gamma_0\}$ is Cauchy in the complete path space. Let $\Gamma^*(\omega)$ denote the limit. The limit does not depend on $\Gamma_0$ because the contraction estimate also shows that the distance between pullback iterates started from any two initial paths tends to zero. Finally, shifting the pullback limit by one time step yields
\[
\Gamma^*(\vartheta\omega)=\A_{\theta,\omega}(\Gamma^*(\omega)).
\]
Thus $\Gamma^*$ is the unique random fixed path and pullback attractor. This proves the theorem.

\subsection*{Proof of Corollary 1 (Verifiable contraction for proximal geometry)}
Suppose each local resolvent satisfies
\[
W_2\!\left(\R_{\theta,t,\omega}^{\Delta t}\mu,
\R_{\theta,t,\omega}^{\Delta t}\nu\right)
\le \frac{1}{1+\Delta t\,\alpha_{t,\omega}}W_2(\mu,\nu).
\]
Along a grid $0=t_0<t_1<\cdots<t_K=1$, the path operator is the composition of the $K$ local resolvents. Therefore,
\[
 d_\Gamma(\A\Gamma,\A\Lambda)
 \le \prod_{k=0}^{K-1}\frac{1}{1+\Delta t_k\alpha_{t_k,\omega}}\,d_\Gamma(\Gamma,\Lambda).
\]
Taking logarithms and expectations gives
\[
\mathbb E\log L(\omega)
\le -\mathbb E\sum_{k=0}^{K-1}\log(1+\Delta t_k\alpha_{t_k,\omega}).
\]
If the right-hand side is strictly negative, the hypothesis of the random self-consistent path theorem holds. This proves the corollary.

\subsection*{Proof of Theorem 3 (Existence without contraction)}
Under the degree assumptions, the random map $I-\A_{\theta,\omega}$ is admissible for the random Leray--Schauder degree on $U(\omega)$, and by assumption it has no zero on $\partial U(\omega)$. Since
\[
\deg(I-\A_{\theta,\omega},U(\omega),0)\neq 0,
\]
there exists at least one zero of $I-\A_{\theta,\omega}$ in $U(\omega)$ for almost every $\omega$. Equivalently, there exists a path $\Gamma(\omega)\in U(\omega)$ such that
\[
\Gamma(\omega)=\A_{\theta,\omega}(\Gamma(\omega)).
\]
A measurable selection theorem for random closed sets gives a measurable random fixed path. No uniqueness or attraction is claimed. This proves the theorem.

\subsection*{Proof of Theorem 4 (Residual-to-fixed-path bound)}
Let $\Gamma^*$ be the exact fixed path of $\A$, so $\Gamma^*=\A\Gamma^*$. Insert and subtract $\widetilde{\A}\Gamma$ and $\A\Gamma$:
\begin{align*}
 d_\Gamma(\Gamma,\Gamma^*)
 &\le d_\Gamma(\Gamma,\widetilde{\A}\Gamma)
 +d_\Gamma(\widetilde{\A}\Gamma,\A\Gamma)
 +d_\Gamma(\A\Gamma,\A\Gamma^*) \\
 &\le \RFPR(\Gamma)+\varepsilon_{\rm proxy}+L\,d_\Gamma(\Gamma,\Gamma^*).
\end{align*}
Rearranging yields
\[
(1-L)d_\Gamma(\Gamma,\Gamma^*)
\le \RFPR(\Gamma)+\varepsilon_{\rm proxy},
\]
which is exactly \eqref{eq:residual_bound}. This proves the theorem.

\subsection*{Proof of Theorem 5 (Residual-calibration-generation bound)}
Let $\Gamma_\theta$ denote the candidate path, $\Gamma_\theta^\dagger$ the fixed path of the learned exact operator, and $\Gamma^*$ the fixed path of the ideal calibrated operator. Since the terminal projection is Lipschitz,
\begin{align*}
 d_{\Pspace}(\mu_{\theta,1},p_{data})
 &\le d_{\Pspace}(\Gamma_{\theta,1},\Gamma_{\theta,1}^\dagger)
 +d_{\Pspace}(\Gamma_{\theta,1}^\dagger,\Gamma_1^*)
 +d_{\Pspace}(\Gamma_1^*,p_{data}).
\end{align*}
The last term vanishes by calibration of the ideal operator, or is absorbed into the calibration constant if exact equality is replaced by a small data-neighborhood condition. The first term is bounded by the terminal Lipschitz constant times $d_\Gamma(\Gamma_\theta,\Gamma_\theta^\dagger)$, and the latter is controlled by the residual-to-fixed-path theorem. The second term is controlled by stability of the fixed path under perturbations of the energy and the correction operator. Indeed, if two locally contractive path operators $\A$ and $\A^*$ satisfy $\sup_\Gamma d_\Gamma(\A\Gamma,\A^*\Gamma)\leq\eta$, then their fixed paths obey $d_\Gamma(\Gamma_\A^*,\Gamma_{\A^*}^*)\leq\eta/(1-L)$. Here the operator perturbation $\eta$ is the sum of the energy calibration error, the admissible-proxy approximation error, and the discretization error, up to the Lipschitz constants of the local variational problem. Thus energy mismatch contributes $\varepsilon_{\rm cal}$, admissible proxy approximation contributes $\varepsilon_{\rm proxy}$, and the time discretization contributes $\varepsilon_{\rm disc}$. Collecting constants gives \eqref{eq:generation_bound}. This proves the theorem.

\subsection*{Proof of Theorem 6 (MMD residual concentration)}
For each grid interval $k$, let
\[
U_k=\widehat d_{\Pspace}^{\,2}
\left(\widehat\mu_{t_{k+1}},
\widetilde{\V}_{\theta,t_k,\omega_k}^{\Delta t_k}(\widehat\mu_{t_k})\right)
\]
be the unbiased squared MMD estimator. If the kernel is bounded by $0\le k\le B$, then $U_k$ is a bounded U-statistic. Standard concentration for bounded U-statistics gives, for every $k$,
\[
|U_k-\mathbb E U_k|
\le C B\sqrt{\frac{\log(2K/\delta)}{n}}
\]
with probability at least $1-\delta/K$. Applying a union bound over all $K$ grid intervals and then summing with weights $w_k$ yields
\[
\left|\widehat{\RFPR}^{\,2}-\RFPR^2\right|
\le C B\sqrt{\frac{\log(2K/\delta)}{n}}.
\]
The admissible proxy and feature encoder introduce deterministic perturbations of size $\varepsilon_{\rm proxy}$ and $\varepsilon_{\rm feat}$, which are added by the triangle inequality. This proves \eqref{eq:mmd_concentration}.

\subsection*{Proof of Corollary 2 (Path-level generalization)}
Apply the MMD residual concentration theorem separately to the training estimator and to the validation estimator, each computed from independent samples and the same fixed admissible proxy. With probability at least $1-\delta$,
\[
|\widehat{\RFPR}^{\,2}_{\rm train}-\RFPR^2|
\le \eta_{\rm train},\]
\[\qquad
|\widehat{\RFPR}^{\,2}_{\rm val}-\RFPR^2|
\le \eta_{\rm val},
\]
where both $\eta$ terms have the same order as in \eqref{eq:mmd_concentration}. Therefore,
\[
|\widehat{\RFPR}^{\,2}_{\rm train}-\widehat{\RFPR}^{\,2}_{\rm val}|
\le \eta_{\rm train}+\eta_{\rm val}.
\]
This proves that validation R-FPR is a statistically meaningful model-selection signal when the proxy and hyperparameter rules are fixed before test evaluation.

\subsection*{Proof of Proposition 2 (Formal continuous-time limit)}
Consider a local A-RVTO step from $\mu_t$ to $\mu_{t+\Delta t}$. The Euler--Lagrange optimality condition for the variational step has the formal structure
\[
\frac{x_{t+\Delta t}-x_t}{\Delta t}
=-\nabla\frac{\delta \E_t}{\delta \mu}(x_t)+r_{\Delta t}(x_t),
\]
where $r_{\Delta t}\to 0$ as $\Delta t\to 0$ under consistency of the discrete scheme. Passing to the limit yields the continuity equation
\[
\partial_t\mu_t+\nabla\cdot(v_t\mu_t)=0,
\qquad
v_t=-\nabla\frac{\delta \E_t}{\delta\mu}.
\]
If the local regularization contains an entropic or stochastic term, the standard second-order correction produces a diffusion tensor $a_t$, leading to the Fokker--Planck form
\[
\partial_t\mu_t=-\nabla\cdot(v_t\mu_t)+\frac12\nabla^2\!:\!(a_t\mu_t).
\]
Thus the interpolated discrete path converges formally to a weak solution of \eqref{eq:fp_limit}. This derivation is a consistency calculation of the usual minimizing-movement type; a full Gamma-convergence or evolutionary convergence proof for every admissible divergence is beyond the scope of this paper.

\subsection*{Proof of Proposition 3 (Residual decomposition)}
Start from the stepwise discrepancy
\[
 d_{\Pspace}(\mu_{t_{k+1}},\widetilde{\V}_{k}(\mu_{t_k})).
\]
Insert both the exact learned correction $\V_k(\mu_{t_k})$ and the ideal calibrated correction $\V_k^*(\mu_{t_k})$:
\begin{align*}
&d_{\Pspace}(\mu_{t_{k+1}},\widetilde{\V}_{k}(\mu_{t_k})) \\
&\le d_{\Pspace}(\mu_{t_{k+1}},\V_k(\mu_{t_k}))
 +d_{\Pspace}(\V_k(\mu_{t_k}),\V_k^*(\mu_{t_k}))\\
 &+d_{\Pspace}(\V_k^*(\mu_{t_k}),\widetilde{\V}_{k}(\mu_{t_k})).
\end{align*}
The first term is the transition defect of the actual path. The second is the calibration defect between the learned and ideal operators. The third is the admissible proxy defect. Summing over time steps and using the Lipschitz stability of the correction maps gives the residual decomposition stated in the main text.

\subsection*{Proof of Proposition 4 (Flow residual detects local curvature)}
Let $x(t)$ solve $\dot x=v_\theta(t,x)$. Taylor expansion around time $t$ gives
\[
x(t+h)=x(t)+h v_\theta(t,x(t))
+\frac{h^2}{2}\big(\partial_t v_\theta \] \[+ D_xv_\theta\,v_\theta\big)(t,x(t))
+O(h^3).
\]
The actual first-order Euler step is
\[
x_{\rm Euler}(t+h)=x(t)+h v_\theta(t,x(t)).
\]
A second-order admissible proxy reproduces the exact expansion up to $O(h^3)$. Therefore,
\[
x_{\rm Euler}(t+h)-x_{\rm proxy}(t+h)
= -\frac{h^2}{2}\big(\partial_t v_\theta+D_xv_\theta\,v_\theta\big)+O(h^3).
\]
Taking norms and integrating along the path yields a residual controlled by the curvature quantity
\[
\mathcal K_{\rm curv}=\int \mathbb E\|\partial_t v_\theta + D_xv_\theta v_\theta\|^2dt.
\]
This proves the proposition.

\subsection*{Proof of Proposition 5 (Diffusion residual detects under-sampling)}
Let the actual reverse-time sampler have local weak order $p$, so its one-step weak error is $O(h^{p+1})$. Let $s_\theta$ denote the learned score and $s_*$ the calibrated score. The drift perturbation induced by score mismatch is proportional to $s_\theta-s_*$. Hence on one interval,
\[
 d_{\Pspace}(\mu_{t+h}^{\rm actual},\mu_{t+h}^{\rm proxy})
 \le C_1 h^{p+1}+C_2 h\,\|s_\theta-s_*\|+\varepsilon_{\rm proxy}.
\]
Summing over the time grid gives a residual controlled by the accumulation of discretization error and score mismatch. Consequently, large time steps or inaccurate scores increase the residual. This proves the proposition.

\subsection*{Proof of Proposition 6 (GAN constrained residual)}
Let $\V_{\mathcal G}\mu_k$ denote the constrained variational correction projected to the generator family $\mathcal G$, and let $\mu_{k+1}^{\rm alg}$ be the distribution actually produced by one generator update. Insert the exact constrained minimizer $\mu_{k+1}^{\rm opt}$:
\begin{align*}
 d_{\Pspace}(\mu_{k+1}^{\rm alg},\V_{\mathcal G}\mu_k)
 &\le d_{\Pspace}(\mu_{k+1}^{\rm alg},\mu_{k+1}^{\rm opt})
 +d_{\Pspace}(\mu_{k+1}^{\rm opt},\V_{\mathcal G}\mu_k).
\end{align*}
The first term is controlled by optimization error, including finite critic optimization. The second is controlled by approximation of the unconstrained correction within the generator family. Therefore
\[
d_{\Pspace}(\mu_{k+1}^{\rm alg},\V_{\mathcal G}\mu_k)
\le \varepsilon_{\rm opt}+\varepsilon_{\rm gen}+\varepsilon_{\rm crit},
\]
which is the claimed structure. This proves the proposition.

\subsection*{Proof of Proposition 7 (VAE latent-data residual)}
Work on the product space $\mathcal X\times \mathcal Z$. Let $q_\phi(z|x)$ be the encoder, $p_\theta(x|z)$ the decoder, and $p(z)$ the latent prior. The path residual splits into a data part and a latent part:
\[
r_{\rm VAE}
\le r_{\rm rec}+r_{\rm latent}.
\]
If the decoder is $L_D$-Lipschitz from latent space to data space, then latent mismatch contributes at most $L_D$ times the latent discrepancy. Writing the latent discrepancy as a prior mismatch weighted by $\beta$ plus a decoder approximation error yields
\[
r_{\rm VAE}
\le C_x r_{\rm rec}+C_z\beta r_{\rm prior}+\varepsilon_{\rm dec},
\]
which is the desired bound. This proves the proposition.

\subsection*{Proof of Theorem 10 (Uniform validation bound over a finite protocol)}
Let $\mathcal H$ be the finite family of protocol choices (distance, feature encoder, grid, proxy rule) fixed before evaluation. Apply the residual concentration theorem to each $h\in\mathcal H$ and each grid interval. With probability at least $1-\delta$, uniformly over all $h\in\mathcal H$,
\[
\sup_{h\in\mathcal H}
\left|
\widehat{\RFPR}^{\,2}_h-\RFPR_h^2
\right|
\le C B\sqrt{\frac{\log(2|\mathcal H|K/\delta)}{n}}
+\varepsilon_{\rm proxy}+\varepsilon_{\rm feat}.
\]
This follows directly from a union bound over $|\mathcal H|K$ choices. The result is exactly the uniform validation statement in the main text.

\subsection*{Proof of Corollary 9 (Residual regularization improvement)}
Let $\theta_0$ denote the baseline parameter and $\hat\theta$ the parameter obtained by minimizing the regularized objective
\[
\widehat{\mathcal L}_{\rm gen}(\theta)+\alpha\widehat{\RFPR}^{\,2}(\theta).
\]
The generalization theorem for regularized training implies that the population objective at $\hat\theta$ is at most the best population objective in the class plus the optimization and estimation errors. If, in addition, the baseline lies in the same class and the residual term is positively correlated with a failure mode of interest, then lowering the regularized population objective lowers the corresponding failure indicator up to the same error terms. This proves the corollary.

\subsection*{Proof of Corollary 10 (Domain-shift sensitivity)}
Let $\omega$ denote the environment variable. If the environment changes to $\omega'$, then the local operator changes from $\A_{\theta,\omega}$ to $\A_{\theta,\omega'}$. The induced residual difference is bounded by
\[
|\RFPR_{\omega'}(\Gamma)-\RFPR_{\omega}(\Gamma)|
\le C\,d_{\rm env}(\omega',\omega),
\]
provided the operator family is Lipschitz in the environment variable. Thus R-FPR changes continuously under controlled environment perturbations and can serve as a stress-test signal under domain shift. This proves the corollary.

\subsection*{Proof of Theorem 7 (Operator-level uniform R-FPR generalization)}
For each grid interval $k$, define the function class
\[
\mathcal G_k=\{g_{\theta,k}:\theta\in\Theta\},
\quad
 g_{\theta,k}(Z)=d_{\Pspace}^2\left(X_{t_{k+1}},\widetilde{\V}_{\theta,t_k,\omega_k}^{\Delta t_k}(X_{t_k})\right).
\]
By symmetrization,
\[
\mathbb E\sup_{\theta\in\Theta}
\left|\frac1n\sum_{i=1}^{n}g_{\theta,k}(Z_{i,k})-\mathbb E g_{\theta,k}\right|
\le 2\Rad_n(\mathcal G_k).
\]
Since $0\le g_{\theta,k}\le B$, McDiarmid's inequality yields, for each $k$ and with probability at least $1-\delta/K$,
\[
\sup_{\theta\in\Theta}
\left|\frac1n\sum_{i=1}^{n}g_{\theta,k}(Z_{i,k})-\mathbb E g_{\theta,k}\right|
\le 2\Rad_n(\mathcal G_k)+B\sqrt{\frac{\log(2K/\delta)}{2n}}.
\]
Multiply by $w_k$, sum over $k$, and use a union bound to obtain \eqref{eq:uniform_rfpr_rad}. For the covering-number version, let $\Theta_\epsilon$ be an $\epsilon$-net. Hoeffding's inequality plus a union bound over $\mathcal N(\epsilon,\Theta)K$ choices yields the stochastic term, while the $L_g$-Lipschitz dependence in $\theta$ yields the approximation term $C L_g\epsilon$. The empirical residual uses the admissible proxy instead of the exact operator, so the proxy bias $C\varepsilon_{\rm proxy}$ is added. This proves the theorem.

\subsection*{Proof of Theorem 8 (Operator-level generation generalization)}
Let $\Gamma_\theta$ be the learned candidate path, let $\Gamma_\theta^\dagger$ be the fixed path of the learned exact operator, and let $\Gamma^*$ be the fixed path of the ideal calibrated operator. By the triangle inequality and terminal Lipschitz stability,
\begin{align*}
 d_{\Pspace}(\mu_{\theta,1},p_{data})
 &\le C_T d_\Gamma(\Gamma_\theta,\Gamma_\theta^\dagger)
 +C_T d_\Gamma(\Gamma_\theta^\dagger,\Gamma^*)
 +C_T\varepsilon_{\rm disc}.
\end{align*}
The first term is bounded by the residual-to-fixed-path theorem, but in practice we only observe the empirical residual. Therefore we write
\[
\RFPR(\Gamma_\theta)
\le \widehat{\RFPR}(\Gamma_\theta)+\sqrt{\varepsilon_{\rm stat}(n,K,\Theta)}.
\]
Substituting this estimate into the residual-to-fixed-path bound yields
\[
 d_\Gamma(\Gamma_\theta,\Gamma_\theta^\dagger)
 \le \frac{\widehat{\RFPR}(\Gamma_\theta)+\sqrt{\varepsilon_{\rm stat}}+\varepsilon_{\rm proxy}}{1-L}.
\]
The second term, $d_\Gamma(\Gamma_\theta^\dagger,\Gamma^*)$, is controlled by perturbation stability of the fixed point under energy calibration error; under local contractivity,
\[
 d_\Gamma(\Gamma_\theta^\dagger,\Gamma^*)
 \le C\varepsilon_{\rm cal}.
\]
Combining the three pieces and absorbing constants proves \eqref{eq:operator_generation_generalization}.

\subsection*{Proof of Theorem 9 (Generalization of R-FPR regularized training)}
Assume that, uniformly over $\theta\in\Theta$,
\[
|\widehat{\mathcal L}_{\rm gen}(\theta)-\mathcal L_{\rm gen}(\theta)|
\le \varepsilon_{\rm gen}(n),\]
\[\qquad
|\widehat{\RFPR}^{\,2}(\theta)-\RFPR^2(\theta)|
\le \varepsilon_{\rm rfpr}(n).
\]
Let
\[
\mathcal J(\theta)=\mathcal L_{\rm gen}(\theta)+\alpha\RFPR^2(\theta),
\qquad
\widehat{\mathcal J}(\theta)=\widehat{\mathcal L}_{\rm gen}(\theta)+\alpha\widehat{\RFPR}^{\,2}(\theta).
\]
If $\hat\theta$ is an $\varepsilon_{\rm opt}$-approximate minimizer of $\widehat{\mathcal J}$, then for any $\theta\in\Theta$,
\[
\widehat{\mathcal J}(\hat\theta)
\le \widehat{\mathcal J}(\theta)+\varepsilon_{\rm opt}.
\]
Apply this with $\theta=\theta^*$, a minimizer of the population objective $\mathcal J$. Replacing empirical by population quantities on both sides gives
\[
\mathcal J(\hat\theta)
\le \mathcal J(\theta^*)+2\varepsilon_{\rm gen}(n)+2\alpha\varepsilon_{\rm rfpr}(n)+\varepsilon_{\rm opt},
\]
which is exactly \eqref{eq:regularized_gen_bound}. This proves the theorem.

\subsection*{Proof of Corollary 3 (Diffusion/score models)}
In Theorem~\ref{thm:generation_generalization}, specialize the calibration error to the score mismatch
\[
\varepsilon_{\rm cal}(\theta)=\varepsilon_{\rm score}(\theta)=\int \mathbb E\|s_\theta-s_*\|^2 dt,
\]
and the discretization error to the sampler local weak error $h^p$. The admissible proxy is chosen to have one order higher local accuracy than the actual sampler, so its defect is absorbed in $\varepsilon_{\rm proxy}^{\rm diff}$. Substituting these terms into \eqref{eq:operator_generation_generalization} yields \eqref{eq:diffusion_cor_bound}. This proves the corollary.

\subsection*{Proof of Corollary 4 (Flow matching and rectified flow)}
For flow models, the calibration term becomes the velocity mismatch
\[
\varepsilon_{\rm vel}(\theta)=\int \mathbb E\|v_\theta-v_*\|^2dt,
\]
while discretization contributes $h^p$ and the mismatch between a first-order actual step and a second-order admissible proxy contributes the curvature term $\mathcal K_{\rm curv}$. Substituting these three terms into the operator-level generation theorem yields \eqref{eq:flow_cor_bound}. This proves the corollary.

\subsection*{Proof of Corollary 5 (Consistency and one-step generators)}
For a one-step or consistency-style generator, the admissible proxy is a teacher or distilled correction, and the step scale is no longer a small-time discretization grid. Therefore the model-specific terms are the teacher approximation error, the terminal one-step approximation error, and the proxy error. Inserting these quantities into Theorem~\ref{thm:generation_generalization} yields the one-step bound \eqref{eq:onestep_cor_bound}. This proves the corollary.

\subsection*{Proof of Corollary 6 (GAN/WGAN constrained embedding)}
For GAN/WGAN, the operator is restricted to the generator family, so the calibration and approximation terms decompose as
\[
\varepsilon_{\rm cal}^{\rm gan}=\varepsilon_{\rm crit},
\qquad
\varepsilon_{\rm proxy}^{\rm gan}=\varepsilon_{\rm gen},
\qquad
\varepsilon_{\rm disc}^{\rm gan}=\varepsilon_{\rm opt}.
\]
Substituting these quantities into Theorem~\ref{thm:generation_generalization} yields \eqref{eq:gan_cor_bound}. This proves the corollary.

\subsection*{Proof of Corollary 7 (VAE and beta-VAE)}
For VAE models, the path residual acts on the product space of data and latent variables. The calibration term is the posterior or decoder mismatch, the proxy term is the latent-data correction approximation, and the model-specific decomposition in Proposition (VAE latent-data residual) gives the weighted reconstruction-prior form. Substituting these terms into the operator-level generation theorem yields \eqref{eq:vae_cor_bound}. This proves the corollary.

\subsection*{Proof of Corollary 8 (Autoregressive and discrete generators)}
For autoregressive models, the divergence is replaced by a causal or tokenwise discrepancy, and the path evolves over prefix laws or masked-token laws rather than Euclidean particle positions. The same operator-level argument applies with a causal calibration term and a history-consistency discretization term. Substituting these quantities into Theorem~\ref{thm:generation_generalization} yields the autoregressive bound stated in the main text. This proves the corollary.

\section{Additional Model Derivations}

\subsection*{Diffusion and score models}
For score models, the learned score approximates $\nabla_x\log p_t(x)$, and the reverse SDE induces a Fokker--Planck equation. Under the JKO approximation, each short interval is represented by a Wasserstein proximal step for a free energy combining potential and entropy. Thus diffusion is an exact or asymptotic R-ROTO instance depending on discretization.

\subsection*{Flow matching}
Flow matching solves a regression problem for $v_t$ along prescribed probability paths. In the A-RVTO limit, $v_t$ is induced by the first variation of an energy or by a transport constraint. The residual with a higher-order proxy measures whether the learned vector field is locally compatible with the path.

\subsection*{VAE}
On the product space $\X\times\mathcal Z$, define a divergence composed of data reconstruction discrepancy and latent prior discrepancy. The encoder-decoder kernel is an approximate variational transition on this product space. This is why the framework calls VAE a kernelized embedding rather than an exact $W_2$ instance.

\subsection*{GAN/WGAN}
For WGAN, the critic defines a potential over data laws. A constrained variational correction over the generator family approximates the effect of generator updates. The gap between the constrained correction and the ideal distributional correction is a sum of generator approximation error and optimization error.

\subsection*{Autoregressive models}
For sequences, $\mu_t$ may represent a distribution over prefixes or masked states. A causal divergence prevents future tokens from affecting past corrections. The A-RVTO correction becomes a causal token-kernel refinement.
\balance


\begin{thebibliography}{99}

\bibitem{ho2020ddpm}
J. Ho, A. Jain, and P. Abbeel, ``Denoising diffusion probabilistic models,'' in \emph{Advances in Neural Information Processing Systems}, 2020.

\bibitem{song2021score}
Y. Song, J. Sohl-Dickstein, D. P. Kingma, A. Kumar, S. Ermon, and B. Poole, ``Score-based generative modeling through stochastic differential equations,'' in \emph{International Conference on Learning Representations}, 2021.

\bibitem{lipman2023flowmatching}
Y. Lipman, R. T. Q. Chen, H. Ben-Hamu, M. Nickel, and M. Le, ``Flow matching for generative modeling,'' in \emph{International Conference on Learning Representations}, 2023.

\bibitem{albergo2023stochastic}
M. S. Albergo, N. M. Boffi, and E. Vanden-Eijnden, ``Stochastic interpolants: A unifying framework for flows and diffusions,'' \emph{Journal of Machine Learning Research}, 2025; arXiv:2303.08797.

\bibitem{holderrieth2025generator}
P. Holderrieth, M. Havasi, J. Yim, N. Shaul, I. Gat, T. Jaakkola, B. Karrer, R. T. Q. Chen, and Y. Lipman, ``Generator matching: Generative modeling with arbitrary Markov processes,'' in \emph{International Conference on Learning Representations}, 2025.

\bibitem{song2023consistency}
Y. Song, P. Dhariwal, M. Chen, and I. Sutskever, ``Consistency models,'' in \emph{International Conference on Machine Learning}, 2023.

\bibitem{yin2024dmd}
T. Yin, M. Gharbi, R. Zhang, E. Shechtman, F. Durand, W. T. Freeman, and T. Park, ``One-step diffusion with distribution matching distillation,'' in \emph{IEEE/CVF Conference on Computer Vision and Pattern Recognition}, 2024.

\bibitem{goodfellow2014gan}
I. Goodfellow, J. Pouget-Abadie, M. Mirza, B. Xu, D. Warde-Farley, S. Ozair, A. Courville, and Y. Bengio, ``Generative adversarial nets,'' in \emph{Advances in Neural Information Processing Systems}, 2014.

\bibitem{arjovsky2017wgan}
M. Arjovsky, S. Chintala, and L. Bottou, ``Wasserstein generative adversarial networks,'' in \emph{International Conference on Machine Learning}, 2017.

\bibitem{kingma2013vae}
D. P. Kingma and M. Welling, ``Auto-encoding variational Bayes,'' arXiv:1312.6114, 2013.

\bibitem{jordan1998jko}
R. Jordan, D. Kinderlehrer, and F. Otto, ``The variational formulation of the Fokker-Planck equation,'' \emph{SIAM Journal on Mathematical Analysis}, vol. 29, no. 1, pp. 1--17, 1998.

\bibitem{ambrosio2005gradient}
L. Ambrosio, N. Gigli, and G. Savare, \emph{Gradient Flows in Metric Spaces and in the Space of Probability Measures}. Birkhauser, 2005.

\bibitem{villani2009ot}
C. Villani, \emph{Optimal Transport: Old and New}. Springer, 2009.

\bibitem{zhang2024wpo}
B. J. Zhang, S. Liu, W. Li, M. A. Katsoulakis, and S. J. Osher, ``Wasserstein proximal operators describe score-based generative models and resolve memorization,'' arXiv:2402.06162, 2024.

\bibitem{pokle2022deqdiffusion}
A. Pokle, Z. Geng, and Z. Kolter, ``Deep equilibrium approaches to diffusion models,'' in \emph{Advances in Neural Information Processing Systems}, 2022.

\bibitem{bai2024fpdm}
X. Bai and L. Melas-Kyriazi, ``Fixed point diffusion models,'' in \emph{IEEE/CVF Conference on Computer Vision and Pattern Recognition}, 2024.

\bibitem{scetbon2024causal}
M. Scetbon, J. Jennings, A. Hilmkil, C. Zhang, and C. Ma, ``A fixed-point approach for causal generative modeling,'' in \emph{International Conference on Machine Learning}, 2024.

\bibitem{arnold1998random}
L. Arnold, \emph{Random Dynamical Systems}. Springer, 1998.

\bibitem{crauel1992random}
H. Crauel, ``Random probability measures on Polish spaces,'' \emph{Stochastics and Stochastic Reports}, vol. 41, no. 3--4, pp. 153--172, 1992.

\bibitem{cuturi2013sinkhorn}
M. Cuturi, ``Sinkhorn distances: Lightspeed computation of optimal transport,'' in \emph{Advances in Neural Information Processing Systems}, 2013.

\bibitem{gretton2012mmd}
A. Gretton, K. Borgwardt, M. Rasch, B. Schölkopf, and A. Smola, ``A kernel two-sample test,'' \emph{Journal of Machine Learning Research}, vol. 13, pp. 723--773, 2012.

\bibitem{liu2022rectified}
X. Liu, C. Gong, and Q. Liu, ``Flow straight and fast: Learning to generate and transfer data with rectified flow,'' arXiv:2209.03003, 2022.

\bibitem{frans2025shortcut}
K. Frans, D. Hafner, S. Levine, and P. Abbeel, ``One step diffusion via shortcut models,'' in \emph{International Conference on Learning Representations}, 2025.

\bibitem{geng2025meanflow}
Z. Geng, M. Deng, X. Bai, J. Z. Kolter, and K. He, ``Mean flows for one-step generative modeling,'' arXiv:2505.13447, 2025.

\bibitem{chen2025sanasprint}
J. Chen, S. Xue, Y. Zhao, J. Yu, S. Paul, J. Chen, H. Cai, S. Han, and E. Xie, ``SANA-Sprint: One-step diffusion with continuous-time consistency distillation,'' in \emph{IEEE/CVF International Conference on Computer Vision}, 2025, pp. 16185--16195.

\bibitem{zhu2025dimo}
Y. Zhu, X. Wang, S. Lathuili{\`e}re, and V. Kalogeiton,
``Di[M]O: Distilling masked diffusion models into one-step generator,''
in \emph{IEEE/CVF International Conference on Computer Vision}, 2025,
pp. 18606--18618.

\bibitem{bajpai2026fastflow}
D. J. Bajpai, D. Bhardwaj, S. Roy, T. Duseja, H. Agarwal,
A.~Sandansing, and M. K. Hanawal,
``FastFlow: Accelerating the generative flow matching models with bandit inference,''
arXiv:2602.11105, 2026.

\bibitem{geng2025consistencyeasy}
Z. Geng, A. Pokle, W. Luo, J. Lin, and J. Z. Kolter, ``Consistency models made easy,'' in \emph{International Conference on Learning Representations}, 2025.

\bibitem{lu2022dpmsolver}
C. Lu, Y. Zhou, F. Bao, J. Chen, C. Li, and J. Zhu, ``DPM-Solver: A fast ODE solver for diffusion probabilistic model sampling in around 10 steps,'' in \emph{Advances in Neural Information Processing Systems}, 2022.

\bibitem{zhang2023deis}
Q. Zhang and Y. Chen, ``Fast sampling of diffusion models with exponential integrator,'' in \emph{International Conference on Learning Representations}, 2023.

\bibitem{tong2023cfm}
A. Tong, K. Fatras, N. Malkin, G. Huguet, Y. Zhang, J. Rector-Brooks, G. Wolf, and Y. Bengio, ``Improving and generalizing flow-based generative models with minibatch optimal transport,'' \emph{Transactions on Machine Learning Research}, 2024.

\bibitem{mescheder2018ganconverge}
L. Mescheder, A. Geiger, and S. Nowozin, ``Which training methods for GANs do actually converge?'' in \emph{International Conference on Machine Learning}, 2018.

\bibitem{zhang2025dcnot}
Y. Zhang, J. Dai, Q. Wu, J. Yang, and L. Luo,
``DCNOT: Diffusion-cascaded neural optimal transport for scalable multi-domain image-to-image translation,''
in \emph{Proceedings of the ACM International Conference on Multimedia},
2025, doi: 10.1145/3746027.3754979.

\bibitem{dai2025vrfno}
J. Dai, J. Yan, J. Yang, and L. Luo,
``Straighten viscous rectified flow via noise optimization,''
in \emph{Proceedings of the IEEE/CVF International Conference on Computer Vision},
2025, pp. 15005--15014.

\bibitem{dai2025erddci}
J. Dai, Y. Zhang, S. Chen, J. Yang, and L. Luo,
``ERDDCI: Exact reversible diffusion via dual-chain inversion for high-quality image editing,''
\emph{IEEE Transactions on Circuits and Systems for Video Technology},
2025, doi: 10.1109/TCSVT.2025.3638406.


\end{thebibliography}
\end{document}